\documentclass[11pt]{article}
\usepackage{fullpage}

\newcommand{\vs}{\vspace{-5pt}}

\usepackage{amsmath,mathtools,amssymb,bm,breakurl,MnSymbol,mathbbol,fmtcount,semtrans,multirow,comment,boldline,algorithm,algorithmic}

\usepackage{tikz,wrapfig,graphicx,epsfig,epsf,caption,subcaption,pgfplots,movie15}
\usetikzlibrary{pgfplots.groupplots}

\usepackage[utf8]{inputenc} 
\usepackage{url}            
\usepackage{booktabs}       
\usepackage{amsfonts}       
\usepackage{nicefrac}       

\usepackage{hyperref,color}
\definecolor{darkred}{RGB}{150,0,0}
\definecolor{darkgreen}{RGB}{0,150,0}
\definecolor{darkblue}{RGB}{0,0,200}
\hypersetup{colorlinks=true, linkcolor=darkred, citecolor=black, urlcolor=darkblue}

\newtheorem{theorem}{Theorem}[section]

\newtheorem{lemma}[theorem]{Lemma}
\newtheorem{corollary}[theorem]{Corollary}

\newtheorem{definition}[theorem]{Definition}

\numberwithin{equation}{section} 

\def \endprf{\hfill {\vrule height6pt width6pt depth0pt}\medskip}
\newenvironment{proof}{\noindent {\bf Proof} }{\endprf\par}

\newcommand{\fc}{f}

\newcommand{\fcl}{f^{\text{lgt}}}
\newcommand{\sft}[1]{\text{sftmx}(#1)}
\newcommand{\Xc}{\mathcal{X}}

\newcommand{\ECE}{\text{ECE}}
\newcommand{\MECE}{\text{Max-ECE}}
\newcommand{\closs}{\Lc_{\text{calib}}}
\newcommand{\NLL}{\text{NLL}}

\newcommand{\tsn}[1]{{\left\vert\kern-0.25ex\left\vert\kern-0.25ex\left\vert #1 
    \right\vert\kern-0.25ex\right\vert\kern-0.25ex\right\vert}}

\newcommand{\eps}{\varepsilon}

\newcommand{\bal}{\boldsymbol{\alpha}}
\newcommand{\st}{\star}
\newcommand{\distas}{\overset{\text{i.i.d.}}{\sim}}

\newcommand{\beq}{\begin{equation}}

\newcommand{\eeq}{\end{equation}}

\newcommand{\nn}{\nonumber}

\newcommand{\eg}{{e.g.}}
\newcommand{\ie}{{i.e.}}

\newcommand{\bd}{\bigodot}

\newcommand{\Lc}{{\cal{L}}}

\newcommand{\Dc}{{\cal{D}}}

\newcommand{\tn}[1]{\|{#1}\|_{\ell_2}}

\newcommand{\Cc}{\mathcal{C}}
\newcommand{\Ac}{\mathcal{A}}

\newcommand{\bbeta}{{\boldsymbol{\beta}}}

\newcommand{\Sc}{\mathcal{S}}

\newcommand{\Mc}{\mathcal{A}}

\newcommand{\vb}{\vct{v}}

\newcommand{\ab}{\vct{a}}
\newcommand{\bb}{\vct{b}}
\newcommand{\ub}{{\vct{u}}}

\newcommand{\Fc}{\mathcal{F}}

\newcommand{\x}{\vct{x}}

\newcommand{\bgl}{{\big |}}

\newcommand{\R}{\mathbb{R}}
\newcommand{\Pro}{\mathbb{P}}

\newcommand{\E}{\operatorname{\mathbb{E}}}

\newcommand{\e}{\mathrm{e}}

\newcommand{\vct}[1]{\bm{#1}}

\newenvironment{myitemize}{\begin{list}{$\bullet$}
		{\setlength{\topsep}{1mm}
			\setlength{\itemsep}{0.25mm}
			\setlength{\parsep}{0.25mm}
			\setlength{\itemindent}{0mm}
			\setlength{\partopsep}{0mm}
			\setlength{\labelwidth}{15mm}
			\setlength{\leftmargin}{4mm}}}{\end{list}}

\begin{document}

\title{On the Role of Dataset Quality and Heterogeneity\\in Model Confidence}

\author{Yuan Zhao$^\dagger$\quad\quad\quad\quad Jiasi Chen$^\dagger$\quad\quad\quad\quad Samet Oymak$^\star$\vspace{15pt}\\ $\dagger$: Department of Computer Science and Engineering~~~~\\$\star$: Department of Electrical and Computer Engineering\\University of California, Riverside}
\maketitle

\begin{abstract}\vspace{-3pt}


Safety-critical applications require machine learning models that output accurate and calibrated probabilities. While uncalibrated deep networks are known to make over-confident predictions, it is unclear how model confidence is impacted by the variations in the data, such as label noise or class size. In this paper, we investigate the role of the dataset quality by studying the impact of dataset size and the label noise on the model confidence. We theoretically explain and experimentally demonstrate that, surprisingly, label noise in the training data leads to under-confident networks, while reduced dataset size leads to over-confident models. We then study the impact of dataset heterogeneity, where data quality varies across classes, on model confidence. We demonstrate that this leads to heterogenous confidence/accuracy behavior in the test data and is poorly handled by the standard calibration algorithms. To overcome this, we propose an intuitive heterogenous calibration technique and show that the proposed approach leads to improved calibration metrics (both average and worst-case errors) on the CIFAR datasets.

%
\end{abstract}

\vspace{-3pt}
\section{Introduction}
\vspace{-3pt}
State-of-the-art classifiers such as deep neural networks typically output probability estimates of class categories.
Particularly in safety-critical applications, these estimates should reflect the true probability of an accurate prediction.
For example, if a collision detection machine learning module on a self-driving car outputs a 0.3 probability of a pedestrian being present, there should indeed be a 30\% chance of a pedestrian being there. Similarly, machine learning has growing importance in healthcare, and classifier confidence is of utmost importance for reliable and safe clinical diagnosis. There is a need for self-aware models that know when they are likely to fail.


Typical neural networks today provide uncalibrated confidence estimates, so various calibration schemes have been proposed in the literature \cite{guo2017calibration,de2018clinically}.
However, such calibration schemes may not perform well with imperfect datasets such as those contaminated with label noise or those with few samples.
Such imperfect datasets can occur in practice; for example, manual labelling by humans could result in errors and noise in the training set, or insufficient data collection from battery-powered sensors may lead to fewer than desired samples for training.
We posit that a classifier trained under such scenarios should be aware of these imperfections and able to output calibrated probabilities reflecting these imperfections.

Furthermore, popular calibration schemes such as temperature scaling \cite{platt1999probabilistic} typically try to find optimal global parameters that are used to calibrate all samples uniformly (\eg~using a single calibration parameter). 
However, complex multi-modal datasets are often heterogeneous and thus a {\em{one size fits all}} approach may be ineffective. For instance, we observe that a classifier may be over-confident in certain classes and under-confident in some others (discussed in Sections \ref{sec theory} and \ref{sec:exp}). Global calibration may fail to treat such heterogeneities, leading to worse calibration performance and uncertainty quantification.
The per-class calibration differences become particularly evident when the individual classes have varying data quality, frequency, and importance \cite{lee2017training,liang2017enhancing,kaiser2017learning}. For example, if the collision detection model has few samples of pedestrian collision but many samples of trash collision, the error of the uncertainty estimate may be high for the pedestrian class, which is undesirable as pedestrian collision is a very important class to predict.


\begin{figure*}
\centering
		\begin{subfigure}[t]{0.28\textwidth}
			\includegraphics[width=\linewidth]{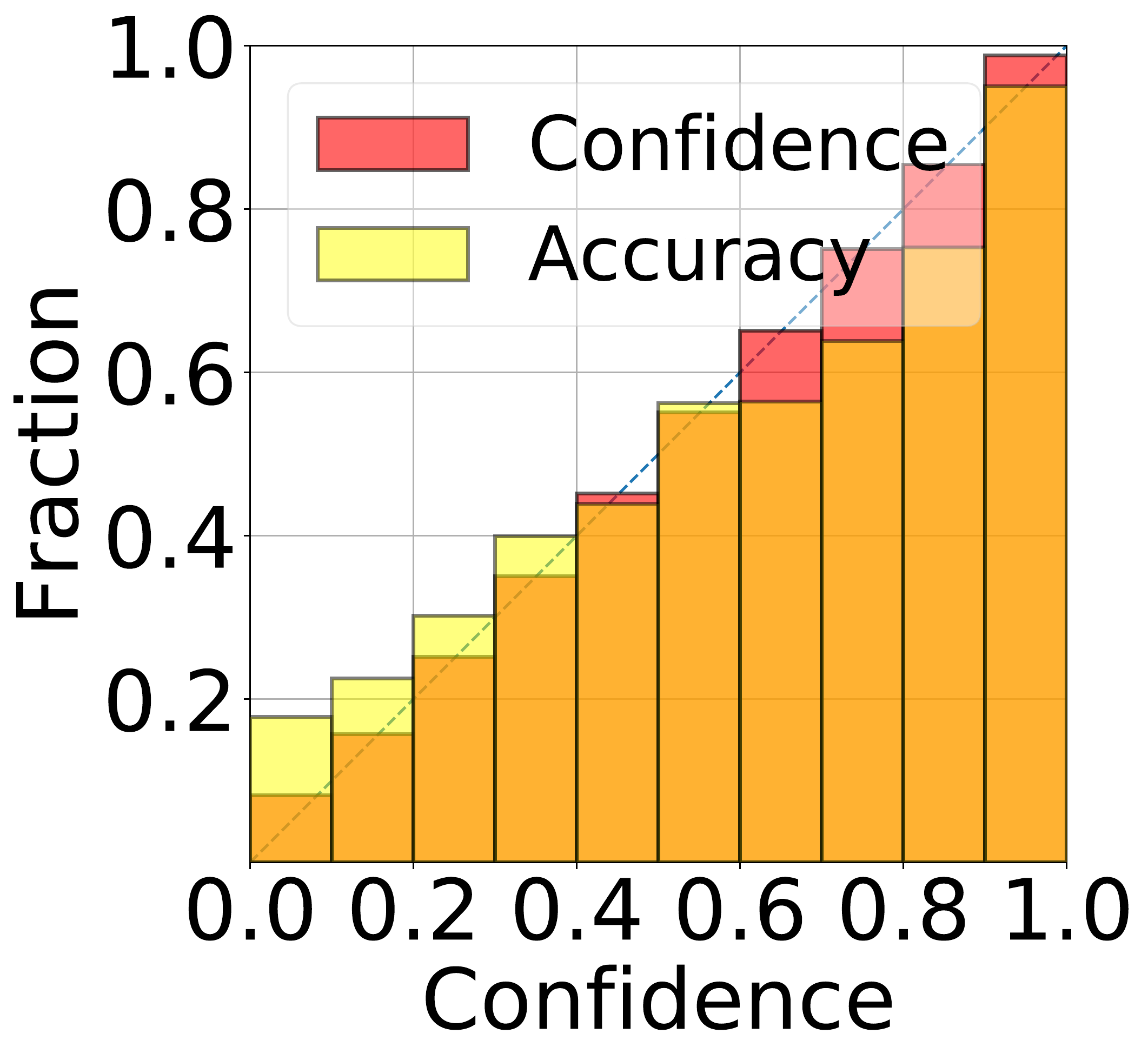}
			\caption{Standard CIFAR-100 model trained with clean data}\label{fig1a}
		\end{subfigure}
		~
		\begin{subfigure}[t]{0.28\textwidth}
			\includegraphics[width=\linewidth]{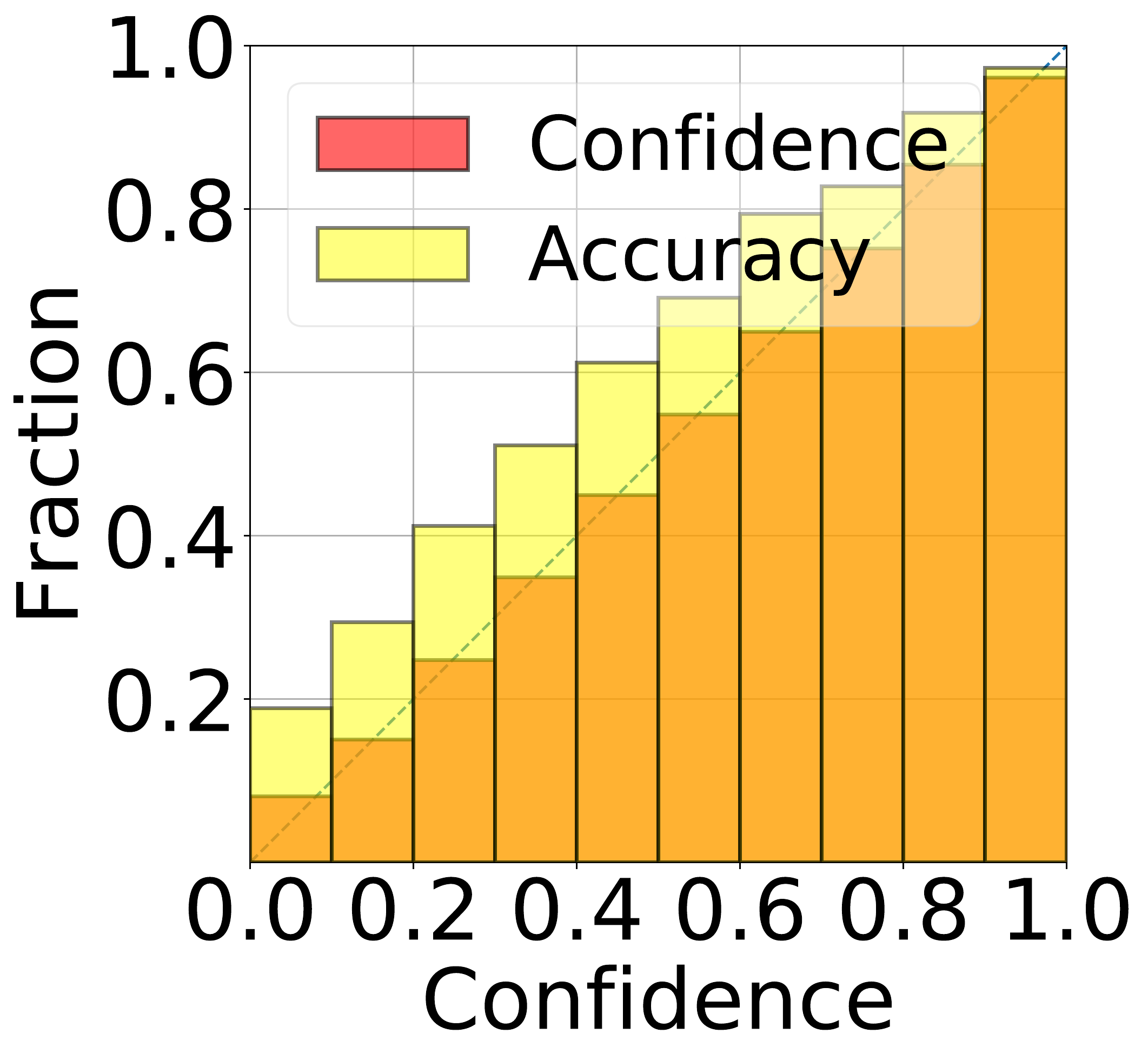}
			\caption{CIFAR-100 model trained with noisy data (30\% chance of label corruption)}\label{fig1b}
		\end{subfigure}
		~
				\begin{subfigure}[t]{0.28\textwidth}
			\includegraphics[width=\linewidth]{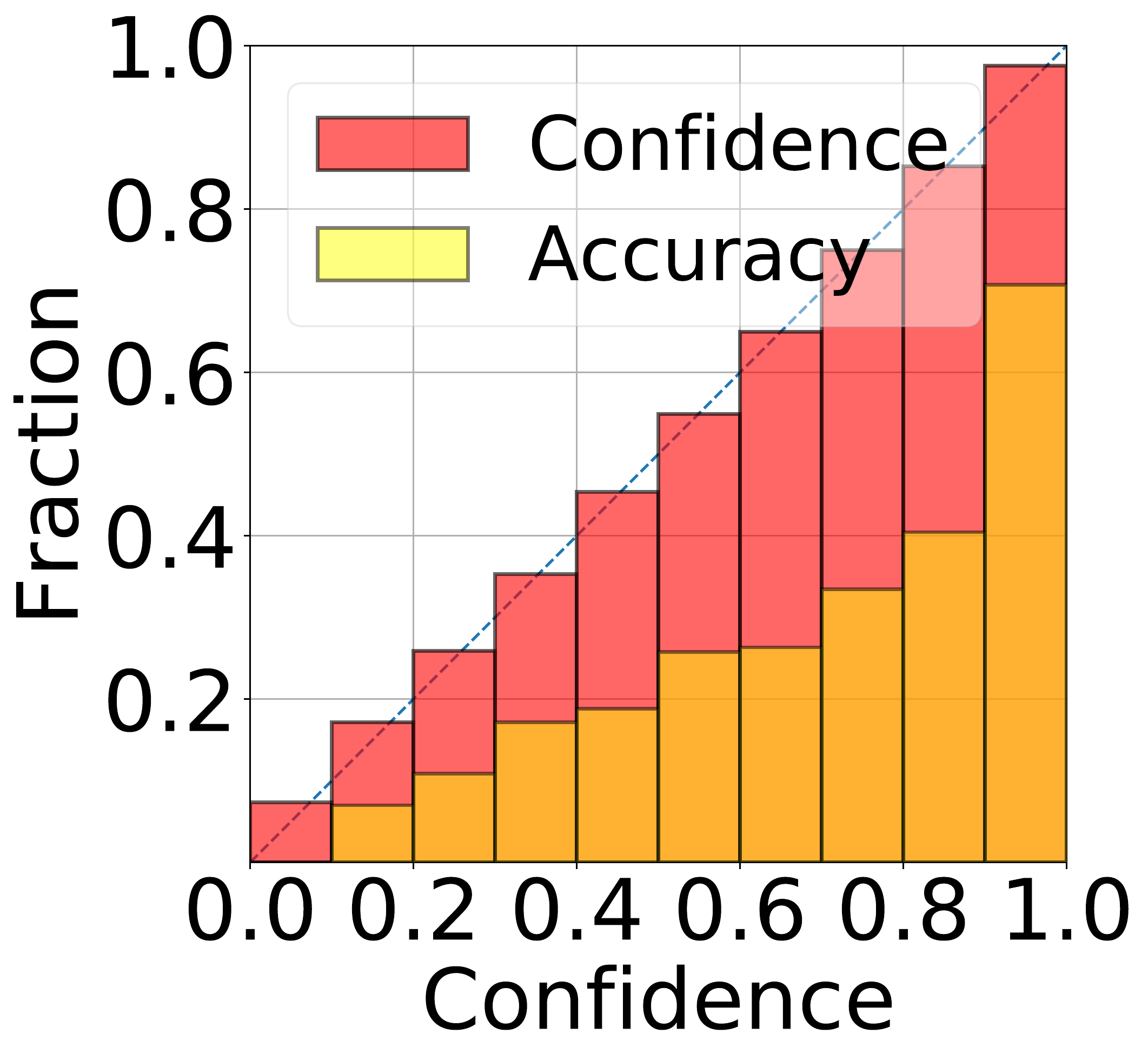}
			\caption{CIFAR-100 model trained with small class sizes (50 labels per class rather than the default 500)}\label{fig1c}
		\end{subfigure}\vspace{0pt}
	\caption{ 
	Reliability diagrams of CIFAR-100, generated by binning the test examples by confidence values, and plotting the average confidence (red), average accuracy (yellow), and their overlap (orange). 
	With perfect calibration, the accuracy should align exactly with the confidence along the diagonal. 
	The model (a) trained with clean data is over-confident (higher confidence than accuracy); (b) trained with noisy data is under-confident (lower confidence than accuracy); and (c) trained with small data size is more over-confident than (a).
	}
	\label{fig:conf_all}\vspace{-0pt}
\end{figure*}

In this work, we explore heterogeneous datasets with varying levels of noise or sample sizes across classes, and investigate per-class calibration both empirically and theoretically. In addition to standard calibration error metrics used in the literature, we also evaluate the calibration success through an intuitive fairness-inspired metric, which captures the worst-case (maximum) calibration error across classes. Our goals are to (a) understand the role of data quality in model confidence via theory and experiments, (b) empirically demonstrate the challenges of calibrating classifiers trained on class-wise heterogenous data, and (c) propose intuitive and effective solutions to correct for this imbalanced calibration. Specifically, our contributions are as follows:

\vspace{0pt}


\vspace{3pt}
\noindent$\bullet$ {\bf{Understanding the role of the data quality.}}\\\emph{Noise imbalance:} We find that label noise in the training data leads to under-confident classifiers (Figure~\ref{fig1b}), and we provide a theoretical justification explaining this observation. This is in contrast, surprisingly, to the over-confidence of deep networks trained with noiseless data previously observed \cite{guo2017calibration,hein2019relu}. \\
\emph{Sample imbalance:} Training sample size similarly has a major effect on classifier confidence. Specifically, in CIFAR100 experiments, a smaller sample size leads to more overconfident classifiers due to lower accuracy (Figure~\ref{fig1c}). 
\vspace{-0pt}

Both of these observations, interestingly, also apply to classifiers trained on heterogenous data. For instance, if the label noise is unbalanced across classes (\eg~some classes have more noise compared to others), the classifier tends to be under-confident on noisier classes and over-confident on cleaner classes (see Figures \ref{fig:noisy reliability} and \ref{fig:undersamp reliability}).

\vspace{3pt}
\noindent$\bullet$ {\bf{Calibration for heterogenous data.}} These observations motivate us to investigate class-wise calibration algorithms. We propose an intuitive and general approach that allows for individually calibrating each class. Specifically, we slice the validation set by predicted class assignments and calibrate each slice separately. Our approach, coupled with temperature scaling method (TS), leads to class-wise temperature scaling (CTS) as a special case. We establish validation sample complexity bounds and demonstrate the benefit of this approach when the classes exhibit noise and sample size imbalances. We also show the benefits of CTS over an alternative approach, vector scaling \cite{guo2017calibration}. 



\vspace{-8pt}
\subsection{Related Works}
\vspace{-5pt}


Uncertainty quantification has a long history in several fields spanning medical diagnosis, meteorology, and risk analysis~\cite{murphy1973new,murphy1977reliability,jiang2011calibrating,platt1999probabilistic}. Due to widespread use of deep learning, there is a growing algorithmic need for training properly calibrated deep networks that are aware of their limitations and can know when they are likely to fail.

It has been empirically observed that modern deep networks tend to be over-confident, \ie~they are more confident than they are accurate, and there is a growing theoretical understanding of this phenomena. Specifically, \cite{du2018gradient2,li2019gradient,bartlett2017spectrally,Belkin:2018ab,Belkin:2018aa} show that large capacity deep networks have the ability to fit any training dataset and achieve 100\% accuracy (a.k.a.~interpolation), leading to very high test confidence~\cite{hein2019relu,guo2017calibration}. Several works 
\cite{guo2017calibration,lakshminarayanan2017simple,liang2017enhancing,de2018clinically,thulasidasan2019mixup,kumar2019verified,hendrycks2019using,naeini,kumar2018trainable} aim to address this calibration challenge. \cite{pleiss2017fairness,hebert2017calibration} relates machine learning fairness to the calibration problem and explores the associated tradeoffs. \cite{snoek2019can} studies model uncertainty under dataset shift and provides large-scale empirical comparison of different calibration techniques. \cite{kull2019beyond} and \cite{kumar2019verified} provides alternative methods improving over Platt scaling.

Calibration is particularly important for detecting rare events and anomalies which is connected to both noisiness and sample imbalance. \cite{lee2017training,liang2017enhancing,devries2018learning} proposes algorithms for confidently discovering out-of-distribution samples by modifying the training process or input samples. In contrast, we investigate the related problem of class imbalances through a generic approach compatible with standard calibration algorithms.




\vspace{-0pt}\section{Problem Setup}\vspace{-2pt}

We consider the supervised classification problem with multiple classes, with an emphasis on deep networks. Denote the joint distribution $\Dc$ of input/output pairs $(X,Y)$ via\vspace{0pt}
\[
P(Y,X)=P(Y\bgl X)P(X).
\]

\vspace{-0pt}
\noindent {\bf{Multiclass setup:}} Input $X\in\Xc$ and output $Y\in\{1,2,\dots,K\}$ are random variables where $Y$ is the true class assignment and $\Xc$ is the input space. $\hat{Y}$ is the predicted class.
Let $\fc:\Xc\rightarrow[0,1]^K$ be a multiclass classifier (e.g.~a deep neural network) mapping an input to a $K$ dimensional vector of class distributions. For an input $X$, $\fc$ outputs a class decision $\hat{Y}=\arg\max_{1\leq k\leq K} \fc(X)_k$ with confidence $\hat{P}=\fc(X)_{\hat{Y}}$, where $f(X)_k$ denotes the $k^\text{th}$ entry of the output vector. $\hat{Y},\hat{P}$ are functions of $f$ and $X$. $\hat{P}_f(X),\hat{Y}_f(X)$ will explicitly highlight this dependence.\vspace{3pt} 

\noindent {\bf{Binary setup:}} For theoretical analysis, we also consider binary classifiers where $Y\in\{0,1\}$ and $f(X):\Xc\rightarrow [0,1]$. We set $\hat{Y}=1$, $\hat{P}=f(X)$ if $f(X)\geq 0.5$ and $\hat{Y}=0$, $\hat{P}=1-f(X)$ if $f(X)< 0.5$. 

Throughout, we assume that the classifier $f$ can be decomposed as a softmax (or logistic) function applied to logits $\fcl$ i.e.~$f(X)=\sft{\fcl(X)}$. This is a natural assumption for modern classifiers such as deep networks. 
We will study how well model confidence at a test sample captures the model accuracy. The mismatch between the model confidence and accuracy is known as the calibration error and there are multiple metrics to assess it \cite{kumar2019verified,guo2017calibration,naeini}. Our approach is agnostic to the specific calibration metric; however, below we will focus on the Expected Calibration Error (ECE). ECE measures the distance between the model accuracy $\Pro(Y=\hat{Y})$ and confidence $\hat{P}$ over fixed confidence levels $\hat{P}=p$ of the predicted label. Its continuous version with respect to the $\ell_1$ metric is given by\vspace{-0pt}
\[
\ECE(f)=\ECE(f,\Dc)=\E_{\hat{P}}[|\Pro(Y=\hat{Y}\bgl \hat{P}=p)-p|].
\]
This continuous version operates in infinitesimal confidence intervals. The discrete version of ECE circumvents this by using binned confidences as defined below.
\begin{definition}[Discrete Expected Calibration Error] \label{ece def}Split the interval $[0,1]$ into $M$ disjoint intervals $(B_i)_{i=1}^M$. The discrete ECE is given by\vspace{-0pt}
\[
\ECE(f)=\sum_{i=1}^M|\Pro(Y=\hat{Y}\bgl \hat{P}\in B_i)-p|\Pro(\hat{P}\in B_i).\vspace{-0pt}
\] 
\end{definition}
In our experiments, the ECE bins are chosen to be equally spaced which is the common approach in the related literature. 
Given a dataset $\Sc=(X_i,Y_i)_{i=1}^n$ with sample size $n$, we denote the empirical (finite sample) version of ECE by $\ECE(f,\Sc)$ obtained by averaging over the dataset.

We will work with the negative log-likelihood (NLL, i.e.~cross-entropy) as the optimization loss function given by $\NLL(f)=-\E[\log(f(X)_{Y})]$. NLL is a popular metric for evaluating uncertainty and is minimized iff $f$ exactly captures the true conditional distribution $P(Y\bgl X)$.  In contrast to ECE, NLL is a continuous function of the samples. This makes it a natural candidate for calibration optimization \cite{guo2017calibration} and we will optimize NLL loss over the validation set for calibration.

%
%
%
\vspace{-0pt}
\section{Theoretical Insights into Model Confidence and Data Quality}\label{sec theory}
Fig.~\ref{fig:conf_all} shows empirically that data quality can greatly affect the model confidence.
Specifically, in Fig.~\ref{fig1b}, a CIFAR-100 model with noisy data (\ie~30\% chance of label corruption) is trained. Compared to the standard CIFAR-100 model with perfect labels (Fig.~\ref{fig1a}), the model with noisy data suffers from under-confidence on the test set.
On the other hand, a CIFAR-100 model trained with small class sizes (only 50 labels per class rather than the standard 500 labels) results in over-confident models (Fig.~\ref{fig1c}), especially compared to the default CIFAR-100 model (Fig.~\ref{fig1a}).
 Towards explaining these observations, in this section we provide theoretical insights on how model uncertainty changes as a function of the noise level and sample size. Our discussion focuses on binary classification with linear classifiers and minimizes binary NLL for training. Specifically, our classifier $f$ will be parameterized by a vector $\ab$ and intercept $b$ via\vspace{-0pt}
\[
f_{\ab,b}(X)=\sft{\ab^TX+b}=\frac{\e^{\ab^TX+b}}{1+\e^{\ab^TX+b}}. \vspace{-0pt}
\]
\vspace{-0pt}
\subsection{The Role of the Sample Size}
\begin{figure}
 	\centering
 	\begin{subfigure}{0.22\textwidth}
 			\includegraphics[width=\textwidth]{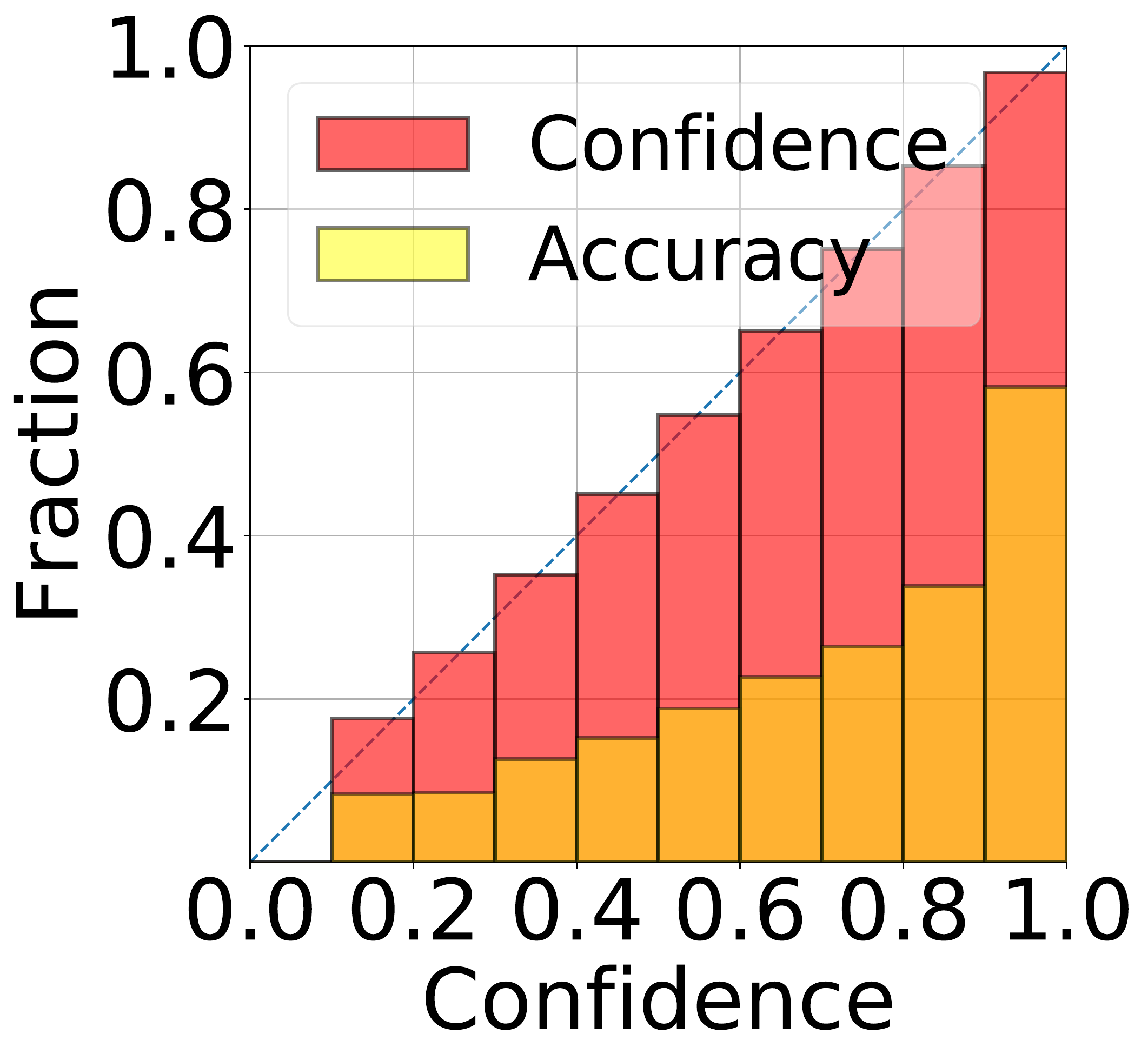}
 			\caption{5\% sampling rate, 65.8\% confidence, 
 			29.5\% accuracy}
 	\end{subfigure}
 	~
 	\begin{subfigure}{0.22\textwidth}
 			\includegraphics[width=\textwidth]{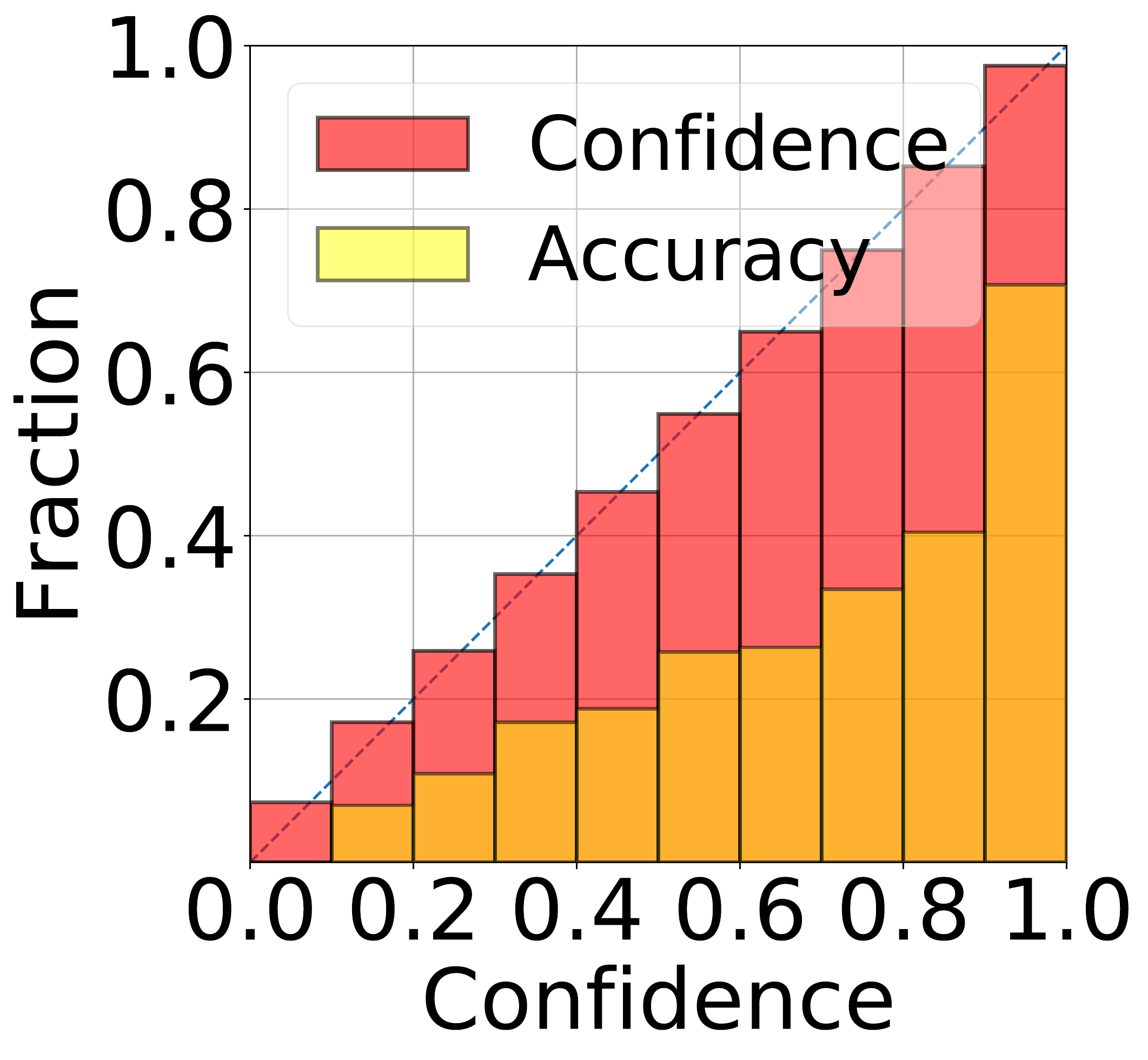}
 			\caption{10\% sampling rate, 72.6\% confidence, 42.2\% accuracy}
 	\end{subfigure}
  	\begin{subfigure}{0.22\textwidth}
 		\includegraphics[width=\textwidth]{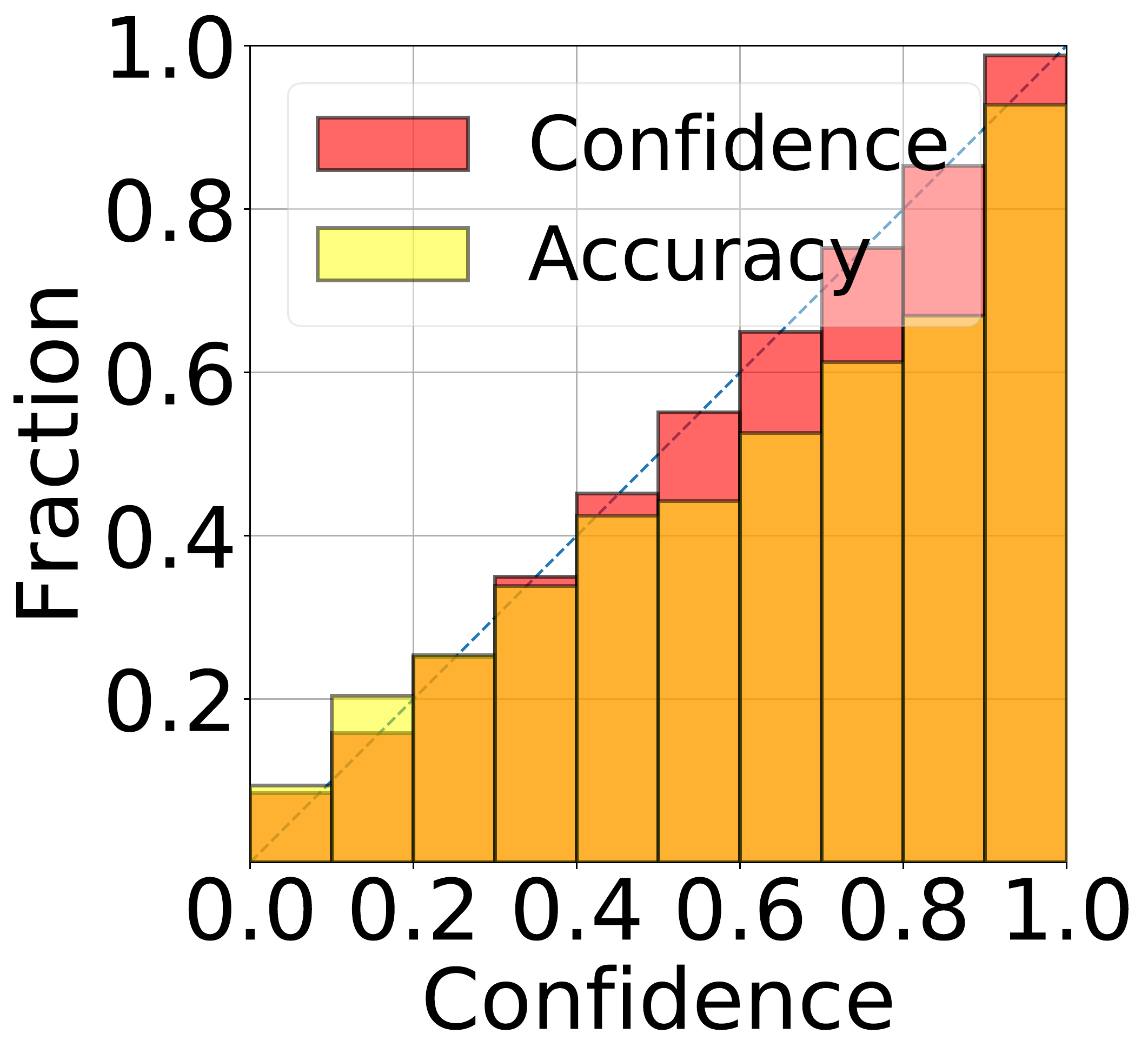}
 		\caption{60\% sampling rate, 83.2\% confidence, 76.3\% accuracy}
 	\end{subfigure}
 	~
 	\begin{subfigure}{0.22\textwidth}
 		\includegraphics[width=\textwidth]{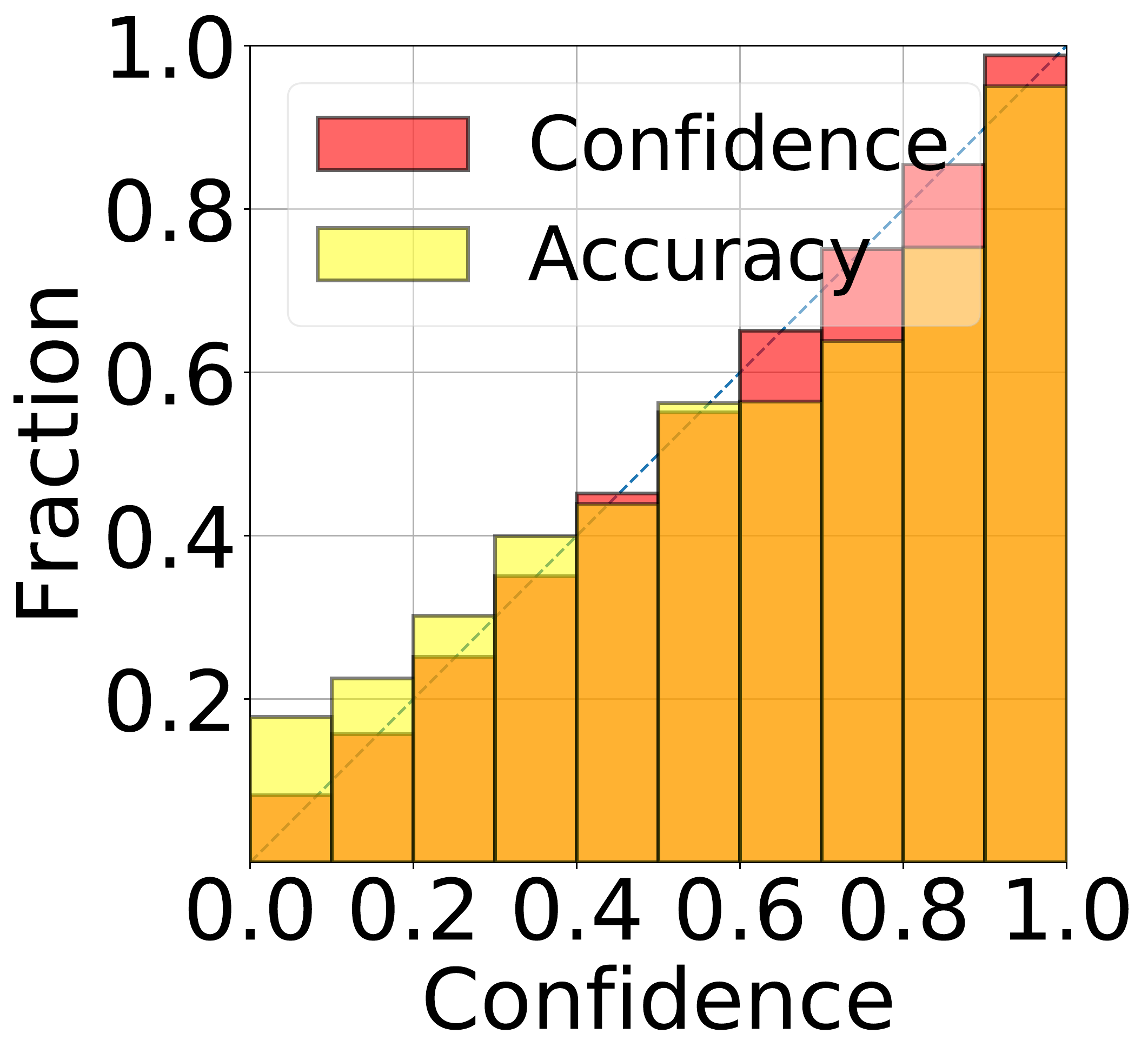}
 		\caption{100\% sampling rate, 84.3\% confidence, 80.9\% accuracy}
 	\end{subfigure}
 	\caption{Reliability diagram and over-confidence for different training dataset sizes for CIFAR-100. A smaller training dataset size leads over-confidence. This is because the model accuracy degrades rapidly with fewer samples. }\label{fig under samp}\vspace{-0pt}
 \end{figure}

First, we explore the role of the dataset size (Fig.~\ref{fig1c}) and 
its impact on model confidence.
Deep networks are often trained until they achieve 100\% training accuracy \cite{zhang2016understanding} which leads to over-confident models. Once a network $f=\sft{\fcl}$ achieves 100\% accuracy, it will still attempt to push NLL to zero. Loss can be pushed to $0$ by scaling up the logits i.e.~by letting $\alpha\rightarrow\infty$ in the class of functions $\sft{\alpha \fcl}$. This eventually leads to classifiers with 100\% confidence in the training data as well as the test data. The reason is that as soon as one entry of $\fcl$ is favorable over the others (which is guaranteed to happen except for degenerate distributions/classifiers), letting $\alpha\rightarrow\infty$ will lead to 100\% prediction confidence. On the other hand, test accuracy highly depends on the sample size and quickly degrades for small datasets. Indeed, the over-confidence in Fig.~\ref{fig1c} compared to Fig.~\ref{fig1a} (also see Fig.~\ref{fig under samp} for details on accuracy/confidence) mostly arises from the lackluster test accuracy. The following result formalizes this observation and states that small sample size can provably lead to further over-confidence.




\begin{theorem}\label{thm under} There exists a distribution $\Dc$ (with unit $\ell_2$ norm input set $\Xc$) as follows. Generate datasets $\Sc_1=(X_i,Y_i)_{i=1}^n\distas\Dc$ and $\Sc_2=(X_i,Y_i)_{i=1}^{30n}\distas\Dc$ and fix $R>0$. Minimize the empirical NLL loss on these datasets to find linear classifiers $f_1,f_2$ as follows.
\[
f_i=\arg\min_{f\in \{f_{\ab,b}~|~\tn{\ab}\leq R\}} \text{NLL}(f,\Sc_i).
\]
Given precision $\eps>0$, choose $R\geq 6\log(50n+\eps^{-1})$. With probability at least $9/10$ (over the proper set $\Sc_1$ or $\Sc_2$), we have the following accuracy and confidence behavior.\vspace{-3pt}
\begin{myitemize}
\item For all inputs $X\in\Xc$ and $i\in\{1,2\}$: $\hat{P}_{f_i}(X)\geq 1-\eps$.
\item $\Pro_{\Dc}(\hat{Y}_{f_1}(X)=Y)\leq 1-\frac{1}{20n}$ and $\Pro_{\Dc}(\hat{Y}_{f_2}(X)=Y)= 1$.
\end{myitemize}\vspace{-3pt}
\end{theorem}
In the setup above, both large dataset ($\Sc_2$) and small dataset ($\Sc_1$) problems lead to arbitrarily high confidence classifiers (over all viable inputs in $\Dc$); however, the model trained on the small dataset is provably less accurate. The proof idea is constructing a distribution where certain features have low probability, thus requiring more data to learn them.

\subsection{The Role of Label Noise}\vspace{-2pt}

For label noise, we work with a noisy dataset model with a discrete distribution over $\Xc=\{\vb,-\vb\}$.\vspace{-3pt} 

\begin{definition} [$\Dc_{\text{noisy}}(p_+,p_-)$] \label{bin data}Fix a vector $\vb\in \R^d$ with unit $\ell_2$ norm and let $\Xc=\{\vb,-\vb\}$. Fix the noise levels $0\leq p_-,p_+\leq 1/2$. Suppose that $\Pro(X=\vb)=1/2$ and the conditional class distributions obey
\[
\Pro(Y=1 \bgl X= \vb)=1-p_+~\text{and}~\Pro(Y=0 \bgl X= -\vb)=1-p_-.
\]
\end{definition}\vspace{-0pt}
The next lemma is a straightforward result that captures the properties of the linear classifier on this noisy data model.
\begin{lemma}\label{thm simple} Fix $1/2> p_+,p_-,p_{\text{test}}\geq 0$. Suppose the data is distributed with $\Dc=\Dc_{\text{noisy}}(p_{\text{test}},p_{\text{test}})$, but the training set is corrupted by label noise in an unbalanced way with distribution $\Dc_{\text{noisy}}(p_+,p_-)$. A linear classifier $f$ minimizing population (infinite sample) training NLL loss obeys\vspace{-3pt}
\begin{align}
&\text{Test confidence over $+$:}~~~~~~\hat{P}(\vb)=f(\vb)=1-p_+\nn\\
&\text{Test confidence over $-$:}~~~~~~\hat{P}(-\vb)=1-f(-\vb)=1-p_-\nn\\
&\text{Test accuracy over either:}~~~\Pro(\hat{Y}=Y\bgl X)=1-p_{\text{test}}.\nn
\end{align}
\end{lemma}\vspace{-0pt}
This lemma highlights that if the training data is noisier than the test (e.g.~$p_+,p_->p_{\text{test}}$), the classifier will be under-confident at test time, explaining the behavior in Fig.~\ref{fig1b}. It also shows that individual classes or inputs can have different confidence levels as a function of noise.

\noindent{\bf{Remarks on under-confidence:}} Recall that once a network achieves $100\%$ training accuracy further training will eventually lead to $100\%$ confidence. Thus large capacity and sufficiently trained networks should not be under-confident. On the other hand, for noisy datasets (e.g.~Fig.~\ref{fig1b}), the training stops before model achieves $100\%$ training accuracy which is the key source of under-confidence\footnote{Maximum validation accuracy is at epoch 61 for Fig.~\ref{fig1b} compared to epoch 172 for Fig.~\ref{fig1a}.}.  
\vspace{-10pt}
\section{Approach for Heterogenous Calibration}\vspace{-5pt}
\label{sec:approach}
\begin{algorithm}[t]
	\caption{Class-wise Calibration}\label{algo:classwise}
	{\bf{Inputs:}} Classifier $f$, validation dataset $\Sc$, regularization $\Gamma$
	
	\hspace{44pt}Calibration loss function~$\closs(\cdot)$ (e.g.~NLL, ECE), search space $\Fc_{\text{cal}}$ (e.g.~Platt scalings)
	
	
	{\bf{Outputs:}} Calibrated classifier $f_{\text{cal}}$
	
	\begin{algorithmic}
		\STATE $\Sc_k=\{(X,Y)\in\Sc~\bgl~ \hat{Y}=k\}$, $\forall~1\leq k\leq K$.
		\STATE Solve the calibration optimization\vspace{-8pt}
		\begin{align}\tag{CC}
		\Cc^\star_k=\min_{\Cc_k}~&\sum_{i=1}^K\closs(\Cc_k(f),\Sc_k)\quad\text{s.t.}\label{MTL}\\
		&\|\Cc_k-\Cc_0\|\leq \Gamma,~\Cc_k\in \Fc_{\text{cal}}~~~\forall~0\leq k\leq K.\nn
		\end{align}
		
		\STATE For a \text{fresh input sample} $X$, $f_{\text{cal}}$ returns $f_{\text{cal}}(X)=\Cc^\star_{\hat{Y}}(f(X))$
	\end{algorithmic}
\end{algorithm}
To address the above challenges of sample size and label noise resulting in different confidence levels, we propose a new calibration method for heterogeneous datasets.
Our approach is summarized in Alg.~\ref{algo:classwise} and applies post-processing on a given classifier $f$. It can use an arbitrary calibration function $\Cc$ (chosen from a set $\Fc_{\text{cal}}$) which takes a classifier $f$ and outputs a calibrated classifier $\Cc(f)$ (e.g.~$\Cc$ applies Platt scaling on $f$) . The core idea is splitting a heterogenous dataset $\Sc$ into homogenous subsets so that $\Cc$ can calibrate each subset individually. The appropriate splitting is a function of the dataset (i.e.~its size and type of heterogeneity), and prior information can guide the subset selection. {A good example is related to fair machine learning where a dataset may be heterogenous with respect to a sensitive input feature (e.g.~race, sex) \cite{pleiss2017fairness}. We can create the sub-datasets, (e.g.~corresponding to different demographic groups) based on the distinct values of the sensitive feature.} 
While our approach can apply to any general splitting policy, in this work, we restrict our attention to the heterogeneity across different classes and focus on class-wise splitting to address unbalanced class distributions.

Specifically, $\Sc$ is split into $K$ subsets $(\Sc_k)_{k=1}^K$ where $\Sc_k$ is the set of samples whose predicted labels $\hat{Y}$ are class $k$. Note that, we use {\em{predicted labels}} for calibration rather than the actual labels, because at the time of inference, we won't have access to the labels and have to infer them.

Our algorithm takes a calibration loss (e.g.~NLL, ECE) and solves the Class-wise Calibration problem \eqref{MTL}. The key idea is individually calibrating each class to obtain $\Cc_k^\st(f)$ from the base function $f$. \eqref{MTL} admits a regularization parameter $\Gamma$ which quantifies the level of multi-task learning. $\Gamma=0$ reduces to standard (non-class-wise) calibration whereas $\Gamma=\infty$ means each class is calibrated by themselves which may be more prone to over-fitting. Finally, for inference in test time, the final calibrated classifier $f_{\text{cal}}$ calls the sub-classifier $\Cc_k^\st(f)$ whenever the predicted tag is class $k$. \vspace{-2pt}

\noindent{\bf{Consistency:}} Suppose $\Cc$ preserves the decision of the classifier i.e.~for all inputs $X$, we have that $\hat{Y}_f(X)=\hat{Y}_{\Cc(f)}(X)$. Then, $f_{\text{cal}}$ is also consistent with $f$ since it is consistent with $\Cc^\st_k(f)$'s. Thus proposed method doesn't affect the classifier decision and it is guaranteed to preserve the accuracy of $f$.

\vs\vspace{-2pt}
\paragraph{Maximum ECE (Max-ECE):} As discussed in Section \ref{sec:exp}, class-wise calibration improves the calibration error suffered by the worst class. We quantify this via the class-conditional distributions $\Dc_k=P(Y,X\bgl \hat{Y}=k)$ and exploring the maximum ECE over all predicted classes defined as\vspace{-1pt}
\begin{align}
\MECE(f)=\max_{1\leq k\leq K} \ECE(f,\Dc_k)\label{max ece}
\end{align}

\subsection{Class-wise Temperature Scaling (CTS)}
Temperature scaling (TS) is a common and successful calibration technique, and is a special case of Platt scaling. TS simply tunes the logits with a single scalar $\alpha$. Thus, given a range $[\alpha_-,\alpha_+]$, TS searches over a function space $\Fc$ defined as\vspace{-0pt}
\begin{align*}
&\Fc=\{f_{\alpha}\quad\text{where}\quad \alpha \in [\alpha_-,\alpha_+]\}\quad\text{and}\quad f_{\alpha}(X)=\sft{\alpha\fcl(X)},
\end{align*}
that minimizes the calibration error over a validation set $\Sc=(Y_i,X_i)_{i=1}^n$. Applying Algorithm \ref{algo:classwise} on TS leads to the Class-wise Temperature Scaling algorithm which applies TS to each class. 
Given hyper-parameters $\alpha_-,\alpha_+,\Gamma$, CTS aims to finds parameters $\bal=(\alpha_i)_{i=0}^K$ over the constraint set
\[
\Mc=\{\bal~\bgl ~|\alpha_j-\alpha_0|\leq \Gamma\quad\text{and}\quad \alpha_-\leq \alpha_0\leq\alpha_+ \quad \text{for all}~1\leq j\leq K\},
\]
where $\Gamma>0$ is the multi-task regularization and $\alpha_-\leq \alpha_+$ are the bounds on the shared variable $\alpha_0$. The $k$'th class gets the temperature scaling $\alpha_k$. We minimize the following calibration loss over the validation dataset $\Sc$ and constraint set $\Mc$
\begin{align}
&\bal^\star=\arg\min_{\bal\in\Mc}\closs(\bal,\Sc)\quad\text{where}\quad \closs(\bal,\Sc)=\frac{1}{n}\sum_{i=1}^n\closs(Y_i,f_{\alpha_{\hat{Y}_i}}(X_i)) .\label{MTCTS}
\end{align}
Here $\alpha_{\hat{Y}}$ means that calibration is done with respect to the predicted label $\hat{Y}$ where the calibrated function is $f_{\alpha}=\sft{\alpha \fcl}$. The following theorem provides a finite sample generalization guarantee for the CTS optimization \eqref{MTCTS} as a function of problem variables.
\begin{theorem}\label{multitask thm} $\fcl$ is a bounded function over $\Xc$ so that $\tn{\fcl(\x)}\leq L_f$. Let $\closs(\cdot,\cdot):\R^K\times \R^K\rightarrow [0,1]$ be a per-sample calibration loss which is $L_c$ Lipschitz in its second variable.  Suppose (for simplicity) that $\alpha_+-\alpha_-\geq 2\Gamma$. Assume we have $n$ i.i.d.~validation samples $\Sc=(X_i,Y_i)_{i=1}^n\distas\Dc$ and we solve the CTS calibration problem \eqref{MTCTS}. With probability at least $1-\delta$, the minimizer of empirical risk $\bal^\star$ achieves small calibration error on test data (i.e.~on the distribution $\Dc$) as follows
\begin{align}
\closs(\bal^\star,\Dc)\leq &\min_{\bal\in\Mc}\closs(\bal,\Dc)+8\sqrt{\frac{\log(2/\delta)}{n}}+\begin{cases}\frac{4}{\sqrt{n}}\sqrt{{\log(1+4(\alpha_+-\alpha_-)\sqrt{n}L_fL_c)}}\quad\text{if}\quad \Gamma\leq \frac{1}{8\sqrt{n}L_fL_c}\\\frac{4}{\sqrt{n}}\sqrt{{\log(\frac{\alpha_+-\alpha_-}{2\Gamma})+(K+1)\log(8\Gamma\sqrt{n}L_fL_c)}}\quad\text{else}
\end{cases}\nn
\end{align}
\end{theorem}
This result shows the finite sample convergence of CTS for bounded calibration losses and Lipschitz functions. When $\Gamma$ is small ($\Gamma\leq 8/\sqrt{n}L_fL_c$), the required sample complexity grows as $n\gtrsim \log(\alpha_+-\alpha_-)+\log L_f+\log L_c$. This quantity is only logarithmic in problem variables and is independent of the number of classes. This is the regime where shared parameter $\alpha_0$ is dominant over the class-specific residuals $(\alpha_i-\alpha_0)_{i=1}^K$. 

When $\Gamma$ is large, sample complexity grows as $n\gtrsim  \log(\alpha_+-\alpha_-)+K(\log \Gamma+\log L_f+\log L_c)$. This is the regime where classes have more flexibility and sample size is linear in $K$. In this regime, except $K$, the dependence on the problem parameters are are again logarithmic. Intuitively the linear dependence on $K$ is not avoidable since there are $K$ variables in the problem. Note that in both scenarios, calibration error increases gracefully in constraints $\Gamma$ and $\alpha_+-\alpha_-$.

It is also informative to understand what happens when each class is optimized individually i.e.~when each $\alpha_k$ can freely minimize the class-conditional calibration error. This corresponds to setting the regularization $\Gamma=\infty$. The following corollary provides bounds for this setup by solving the calibration problem where $\closs$ is same as \eqref{MTCTS}
\begin{align}
\bal^\star=\arg\min_{\bal\in\Mc'}\closs(\bal,\Sc)\quad\text{where}\quad\Mc'=\{\bal~\bgl ~ \alpha_-\leq \alpha_j\leq\alpha_+ \quad \text{for all}~1\leq j\leq K\}.\label{ind prob}
\end{align}
This is identical to individually minimizing calibration error over the sub-datasets $\Sc_k=\{(X_i,Y_i)\in\Sc~\bgl~\hat{Y}_i=k\}$ by fitting the corresponding $\alpha_k$'s.
\begin{corollary} \label{multitask cor}Consider the setup of Theorem \ref{multitask thm} and solve the calibration problem \eqref{ind prob} over the validation set $\Sc$. Define class-conditional distributions $\Dc_k=P(Y,X\bgl \hat{Y}=k)$ and the minimum class probability $p_{\min}=\min_{1\leq k\leq K}\Pro(\hat{Y}=k)$. With probability at least $1-K(\delta+\e^{-p_{\min}n/8})$, the following generalization guarantees hold for all $1\leq k\leq K$
\begin{align}\label{ind bound}
\closs(\alpha_k^\star,\Dc_k)\leq \min_{\alpha_-\leq \alpha\leq \alpha_+}\closs(\alpha,\Dc_k)+6\sqrt{\frac{\log(1+4(\alpha_+-\alpha_-)\sqrt{n}L_fL_c)}{p_{\min}n}}+12\sqrt{\frac{\log(2/\delta)}{p_{\min}n}}
\end{align}
\end{corollary}
This result establishes test calibration error performance for class-wise distributions $\Dc_k$ and associated parameters $\alpha_k$ and follows from a standard union bound argument. Assuming $p_{\min}\propto 1/K$, observe that the calibration error grows as $\sqrt{K/n}$ in a similar fashion to the large $\Gamma$ regime of Theorem \ref{multitask thm}. This is intuitive as each class is calibrated separately i.e.~effective sample size for calibration is $n/K$. Recalling \eqref{max ece} and setting $\closs$ to ECE, this corollary also bounds Max-ECE.

Finally, recall that Algorithm \ref{algo:classwise} and CTS mitigates the challenge of small sample size per class via multi-task regularization. An alternative approach for addressing small class sizes would be grouping classes with similar statistics (e.g.~by noise level, size, etc) together to obtain larger class bundles with homogenous statistics. Once each bundle contains enough samples, we can safely fit individual calibration models (e.g.~temperature scales).\vspace{-5pt}

\paragraph{Connection to vector scaling:} Vector scaling (VS) is a generalization of the temperature scaling and allows for a more refined access to the softmax layer by using $2K$ parameters for calibration (compared to a single parameter in TS). However this may lead to overfitting in the calibration process \cite{guo2017calibration}. Specifically, VS calibrates over a larger class of functions given by 
\begin{align}
f_{\ab,\bb}=\sft{\ab \bd\fcl(X) + \bb}. \label{vecscale}
\end{align} Here $\bd$ is the entrywise product and $f_{\ab,\bb}$ is parameterized by the $K$ dimensional scaling vector $\ab$ and bias vector $\bb$. 

Observe that, ignoring the bias $\bb$, VS is related to CTS, as the entries of $\ab$ provide knobs on how much we wish to emphasize a particular class.
Experimentally, we find that VS is indeed a viable approach for addressing class imbalances. On the other hand, unlike CTS, VS may lead to inconsistency in that the classifier decision may change after calibration. Since each logit is calibrated with its own parameter, the relative order of the calibrated logits may change resulting in different accuracy. We will show that CTS achieves comparable ECE and max-ECE values without compromising the accuracy of the black-box classifier.




\section{Numerical Results}
\label{sec:exp}


\subsection{Simulation Setup}
\label{sec:setup}

\textbf{Datasets:} We use the CIFAR-100 dataset and perform image classification to demonstrate our proposed CTS method (Alg.~\ref{algo:classwise}).
We also experiment with CIFAR-10 (see Table \ref{table:calibration10} and Appendix B), and find similar results to CIFAR-100.
The CIFAR datasets contains 50k training samples and 10k testing samples.
Whenever validation is needed, the original training set is split into 45k training samples and 5k validation samples. We only modify the 45k training samples.
The validation set is unmodified, with equal class sizes of 50 samples each.
All experiments use standard data augmentation techniques of image shifting and horizontal flipping. Experiments are repeated five times with different random seeds. To evaluate the impact of heterogeneous data, we construct two variants of the CIFAR-100 data:
\begin{myitemize}
	\item \emph{Noise-imbalanced dataset construction (Sec.~\ref{sec:noisy}):}
In the training dataset, we add label noise to classes 0 to 49 with the noise rate $\rho$ varying from 0 to 1 (\ie~with $\rho$ probability, a label is randomly assigned to one of the 100 labels). Classes 50 to 99 remained unchanged.
	\item \emph{Size-imbalanced dataset construction (Sec.~\ref{sec:size}):}
	 We under-sample classes 0 to 49 in the training set, with the sampling rate $\rho \in [0.05,1]$. Instead of the usual 450 training samples, under-sampled classes have only $450\rho$ training samples (\eg~$\rho=0.05$ results in the smaller classes containing only 23 samples, producing a highly unbalanced dataset).
	 The final training set includes the 50 under-sampled classes and the other 50 classes.
\end{myitemize}

\noindent\textbf{Comparison algorithms:} We compare the performance of two class-wise approaches (class-wise temperature scaling and vector-scaling) versus two standard approaches that globally apply the same calibration to all samples (temperature scaling and no calibration), as described below. The ECE and max-ECE are computed from the test dataset.

\begin{myitemize}
	\item \emph{Uncalibrated:} The model is not modified after training or calibrated.
	\item \emph{Temperature Scaling (TS):} TS applies the same calibration parameter $\alpha$ to all classes (see Sec.~\ref{sec:approach}).
	\item \emph{Class-wise Temperature Scaling (CTS):} CTS uses a per-class calibration parameter $\alpha_k$ (see Sec.~\ref{sec:approach}, Alg.~\ref{algo:classwise}). We do not apply multitask regularization (i.e.~$\Gamma=\infty$) and instead solve \eqref{ind prob} with $\alpha_-=0,~\alpha_+=\infty$. In implementation, the predicted labels $\hat{Y}$ are obtained from the $\arg\max$ of the softmax probabilities. Then $(\alpha_k)_{k=1}^{100}$ on each sub-dataset $(\Sc_k)_{k=1}^{100}$, where $\hat{Y}=k$, are obtained. 
	\item \emph{Vector Scaling (VS): } For CIFAR-100 data, VS uses 200 parameters to adjust the network's logits as in \eqref{vecscale}. 
\end{myitemize}

\noindent\textbf{Metrics:} We evaluate the 
following metrics:
\begin{myitemize}
	\item \emph{ECE:} For VS and TS, ECE is computed as described in (Def.~\ref{ece def}). For CTS, ECE is computed by merging the calibrated bins: The average confidence and accuracy are calculated from the samples of different predicted labels within bins of the same confidence intervals.   
	\item \emph{max-ECE:} As defined in (\ref{max ece}), for CTS, max-ECE is computed as the maximum ECE across the 100 predicted labels. For TS and VS, max-ECE is computed by binning the already calibrated samples by their predicted labels and selecting the maximum ECE.
\end{myitemize}

\begin{figure}
	\centering
	\begin{subfigure}{0.33\textwidth}
		{\includegraphics[width=\textwidth]{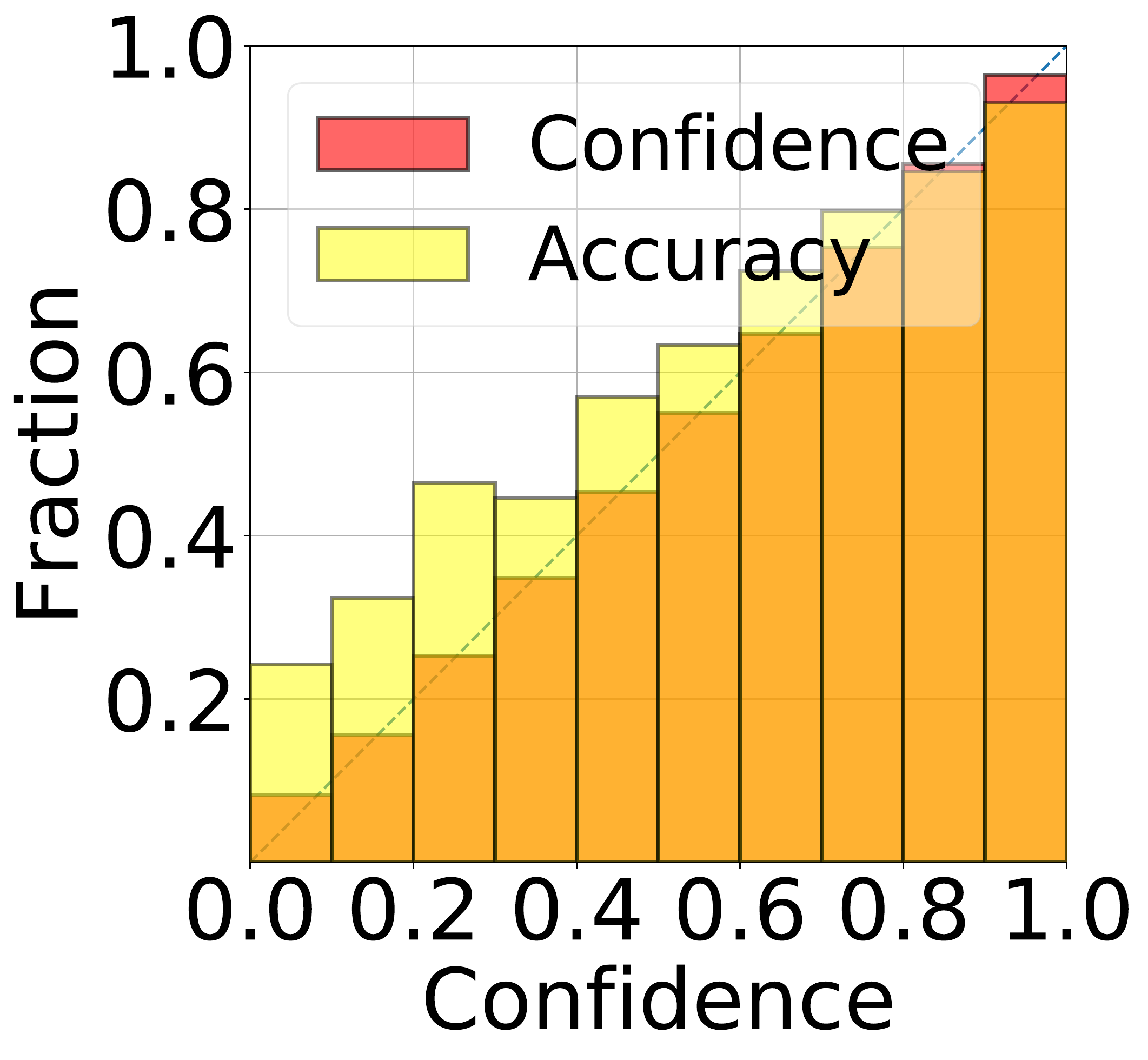}}\vspace{-5pt}
		\caption{Noisy classes}
		\label{fig:noisy reliability1}       
	\end{subfigure}
	\begin{subfigure}{0.33\textwidth}
		{\includegraphics[width=\textwidth]{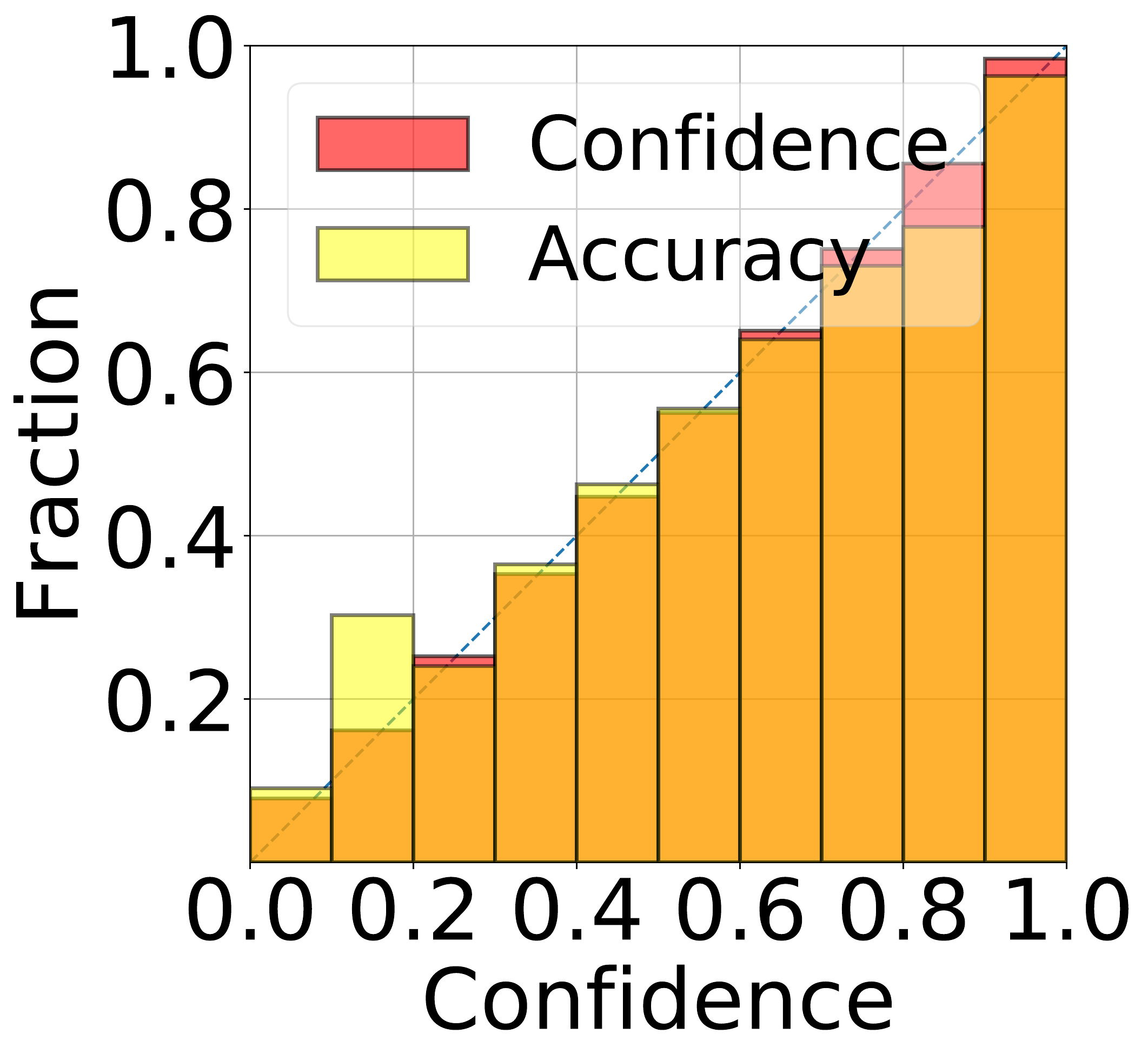}}\vspace{-5pt}
		\caption{Clean classes}
		\label{fig:noisy reliability2}
	\end{subfigure}
	\caption{Reliability diagrams for the noisy and clean subsets of the dataset for noisy CIFAR-100 experiment provided in Section \ref{sec:noisy} (classes 0-49 are 30\% noisy). 
	 }
	 \label{fig:noisy reliability} \vs
\end{figure}

\begin{figure}
	\centering
	\begin{subfigure}{0.33\textwidth}
		{\includegraphics[width=\textwidth]{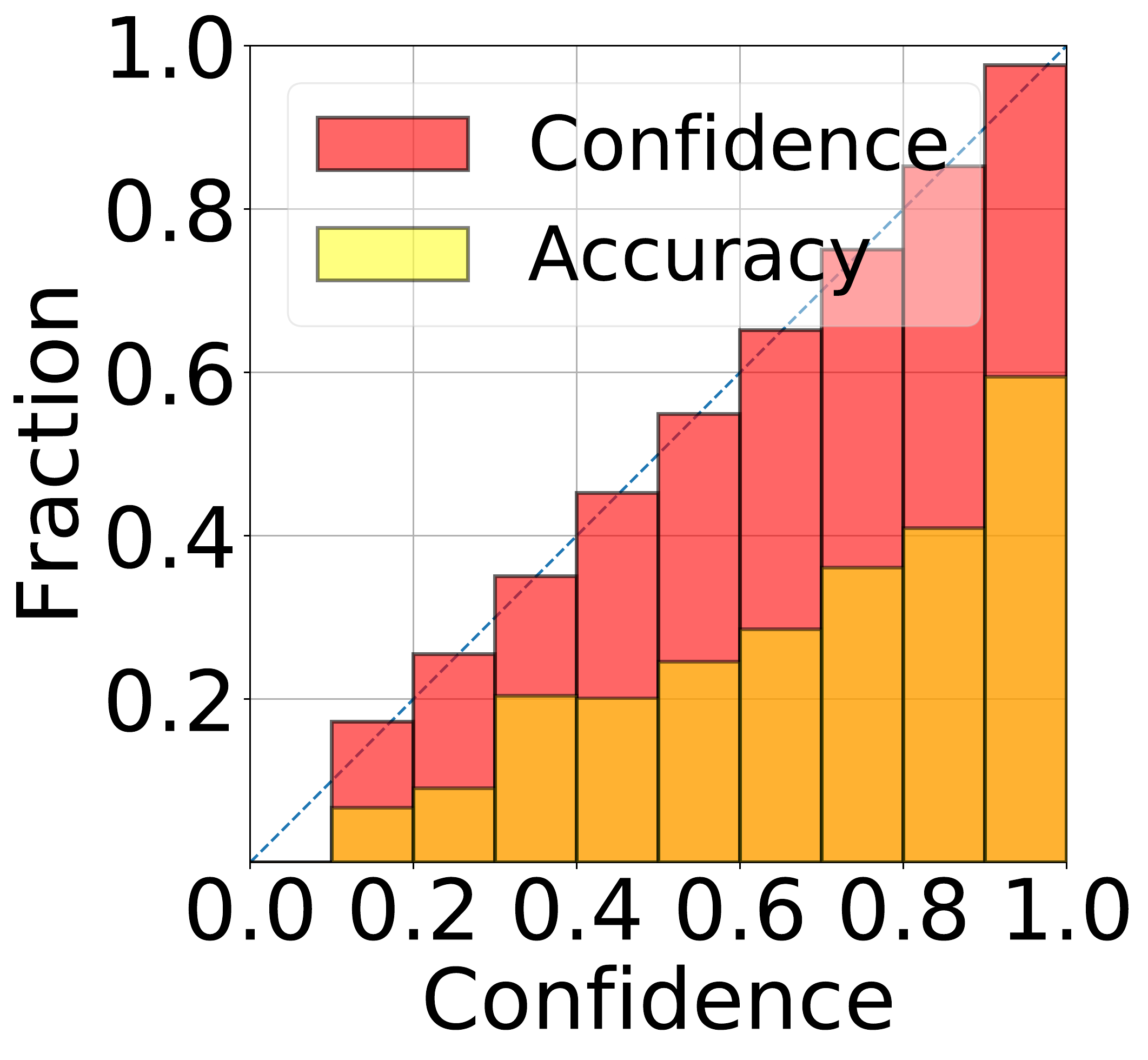}}\vspace{-5pt}
		\caption{Undersampled classes}
		\label{fig:per-class-ece}       
	\end{subfigure}
	\begin{subfigure}{0.33\textwidth}
		{\includegraphics[width=\textwidth]{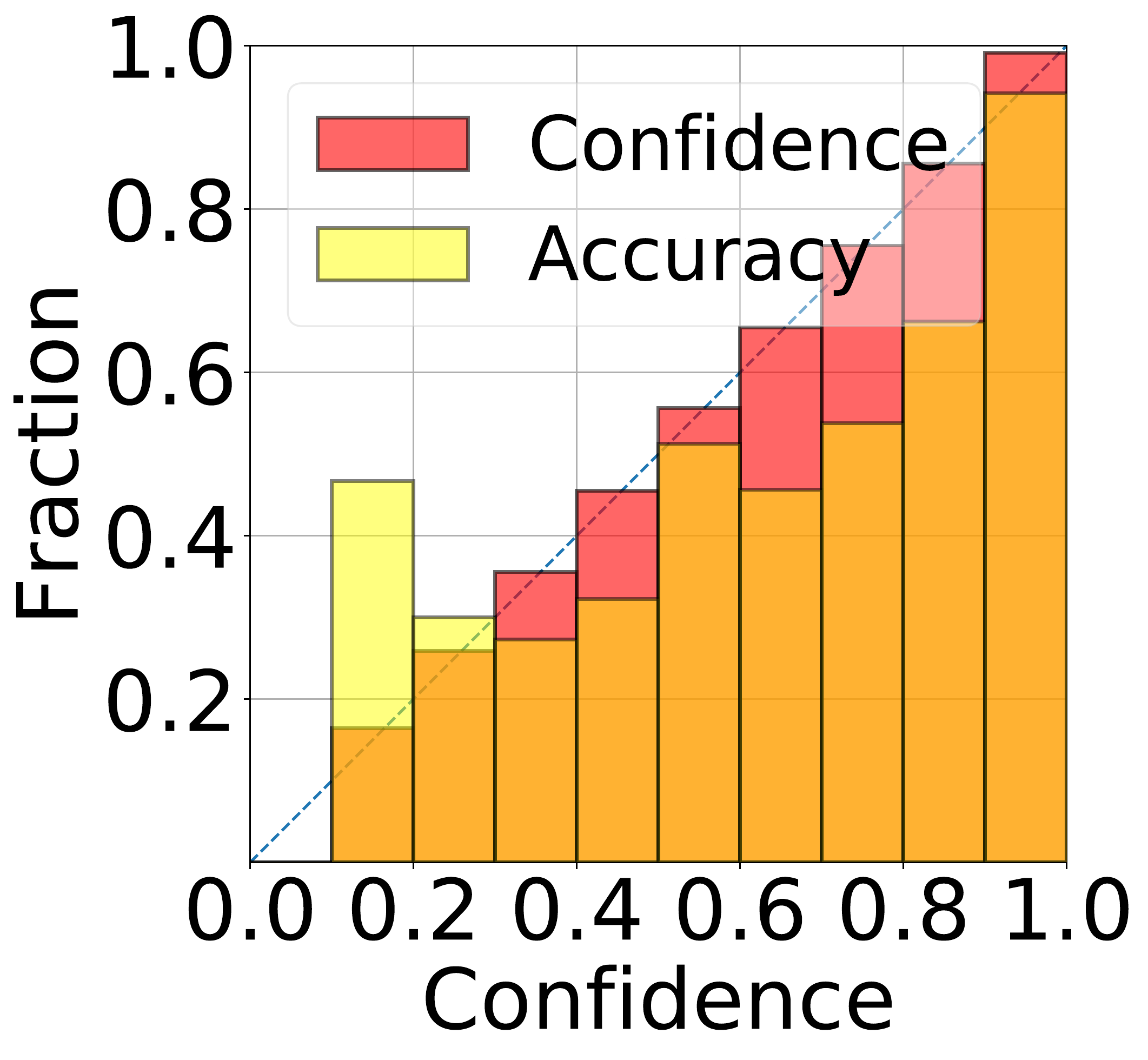}}\vspace{-5pt}
		\caption{Fully sampled classes}
		\label{fig:per-class-accuracy}
	\end{subfigure}
	\caption{Reliability diagrams for the undersampled and clean subsets of the dataset for undersampled CIFAR-100 experiment provided in Section \ref{sec:size} (classes 0-49 are 10\% undersampled). 
	 }
	 \label{fig:undersamp reliability} \vs
\end{figure}

\noindent\textbf{Neural network model and training:} To perform image classification, we utilize the WideResNet-28-10 network model.
We optimize the differentiable NLL loss for calibration optimization (\eg~fitting TS, CTS, VS) as a proxy for ECE and max-ECE. Note that this is a common practice~\cite{guo2017calibration}.
200 training epochs of the WideResNet-28-10 with SGD optimizer were used to fit the data, with cross-entropy as the loss function using Keras and TensorFlow \cite{chollet2015keras,abadi2016tensorflow}. The initial learning rate is initially set to $0.1$ and decreases to $0.02$, $0.004$, $0.0008$ after 60, 120, 160 epochs respectively. 

\subsection{Model Confidence of Heterogenous Datasets before Calibration}
We first study the role of heterogeneity on model confidence at the end of training but before application of any calibration.

\noindent{\bf{Contrasting Noisy vs Clean:}} Our first experiment explores the noise heterogeneity in the training data. We add 30\% label noise on the training data of the CIFAR-100 classes 0-49 and keep the classes 50-99 clean. Figure \ref{fig:noisy reliability} provides separate reliability diagrams for the noisy and clean subsets of the overall CIFAR-100 dataset at the end of training and before any calibration. In consistency with theoretical intuition, this figure demonstrates that noisy classes tend to be under-confident and clean classes tend to be over-confident. The average accuracy over noisy classes 0-49 is 0.689 and average confidence is 0.627. In contrast, average accuracy over clean classes 50-99 is 0.768 and average confidence is 0.781. 

\noindent{\bf{Contrasting Under-sampled vs Full Dataset:}} Our next experiment explores the heterogeneity on the sample sizes within the classes. We under-sample classes 0-49 at 10\% (i.e.~50 per class rather than 500) and classes 50-99 remains untouched. Figure \ref{fig:undersamp reliability} provides reliability diagrams for undersampled vs fully-sampled classes. This figure demonstrates that under-sampled classes tend to be more over-confident than fully-sampled classes. The average accuracy over under-sampled classes 0-49 is 0.396 and average confidence is 0.728. In contrast, average accuracy over fully sampled classes 50-99 is 0.841 and average confidence is 0.909.

These figures indicate that observations of Figure \ref{fig:conf_all} on the effects of noise and undersampling on homogeneous datasets indeed translate to the heterogenous datasets. We emphasize that in Figures \ref{fig:noisy reliability} and \ref{fig:per-class-accuracy}, the subsets of the data that we contrast (clean vs noisy or under-sampled vs fully-sampled) are determined by the actual test labels (rather than the labels predicted by the classifier).

\subsection{Noise-Imbalanced Training Data}
\label{sec:noisy}




\begin{figure}
	\centering
	\begin{subfigure}{0.35\textwidth}
			{\includegraphics[width=\textwidth]{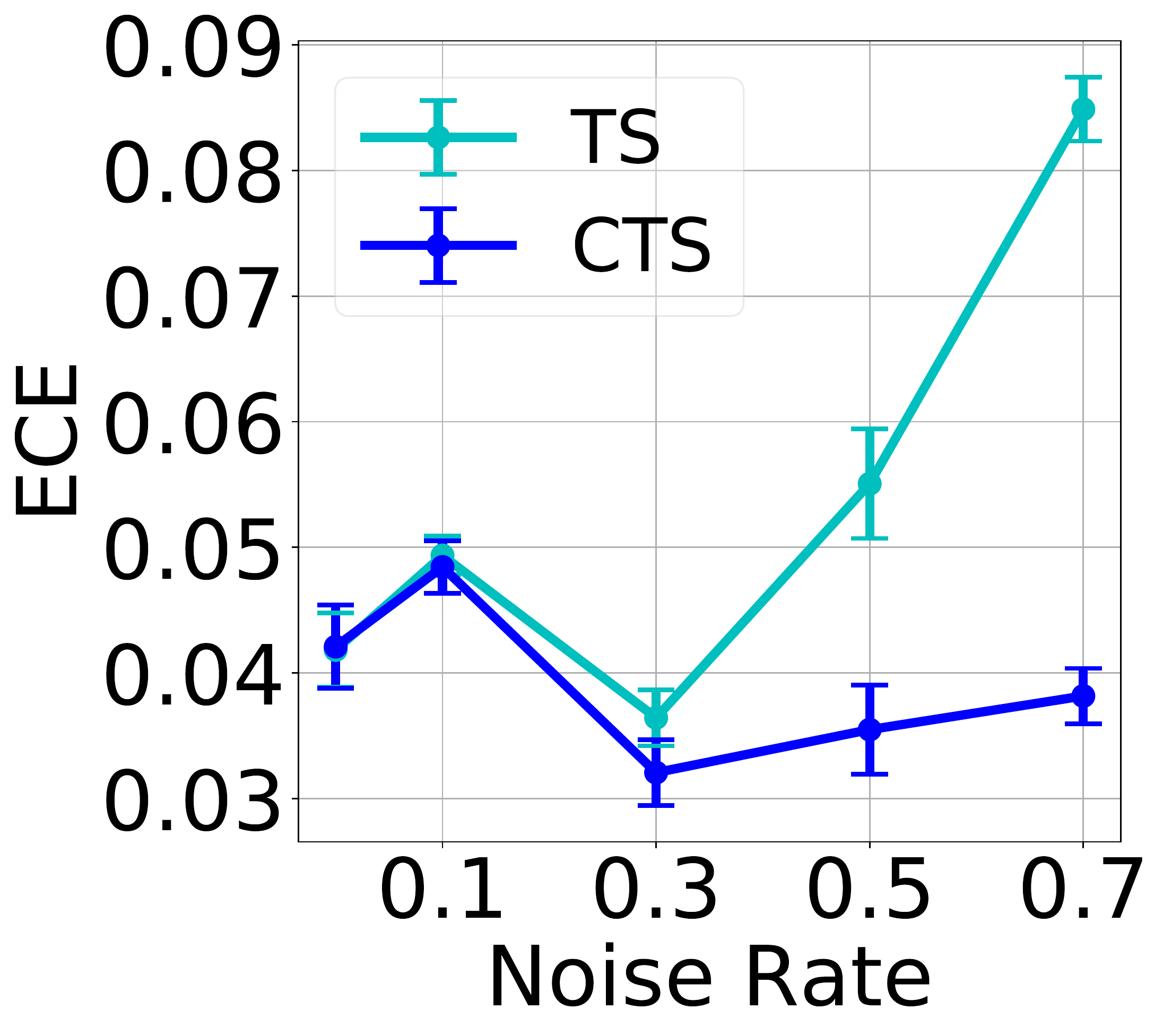}}
			\caption{ECE}       
	\end{subfigure}
	\begin{subfigure}{0.35\textwidth}
			{\includegraphics[width=\textwidth]{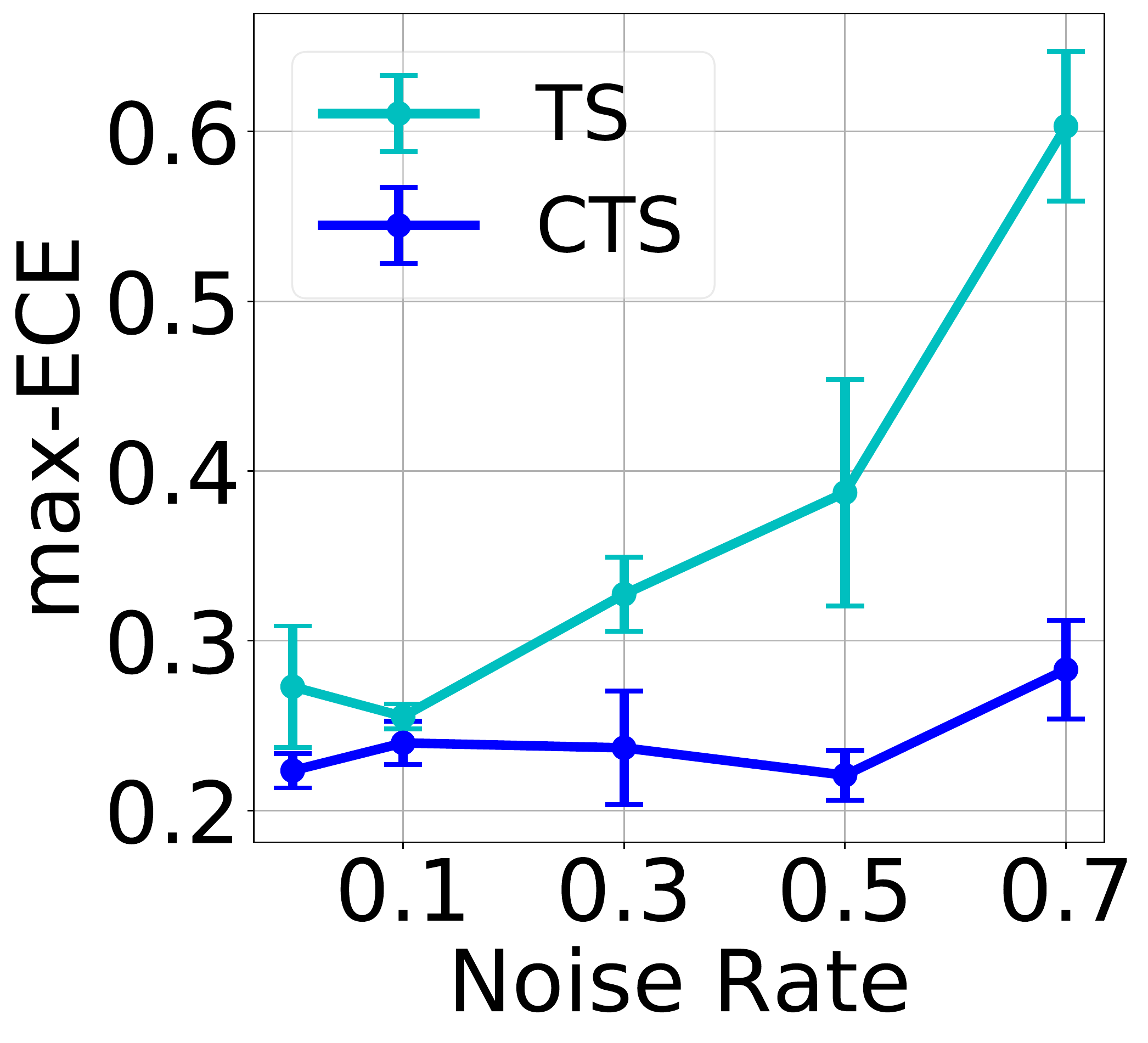}}
			\caption{max-ECE}       
	\end{subfigure}
	\caption{Impact of training data noise on the TS and CTS methods. CTS has lower error in terms of both the ECE and max-ECE metrics.}
	\label{fig:noise}
	
\end{figure}







\textbf{\emph{CTS has lower max-ECE and ECE than TS even with noisy training data.}} We first evaluate the impact of noisy training data on the calibration error. 
In Fig.~\ref{fig:noise}, we plot the ECE and max-ECE as we sweep across different noise rates (as described in Sec.~\ref{sec:setup}).
The CTS method shows a significant improvement (lower ECE and max-ECE) over simple TS, especially when there is more noise in the dataset.
These results suggest that not only CTS can achieve better calibration on individual classes (as shown by the max-ECE plot), but it can also result in a better calibrated model from a global perspective (as shown from the global ECE plot).


\begin{table}
	\begin{center}
		\begin{tabular}{l|c|c|r}
			\toprule 
			\textbf{Alg.} & \textbf{Acc. (\%)} & \textbf{ECE (\%)} & \textbf{max-ECE (\%)}\\
			\midrule 
			Uncal. &  $71.53\pm0.13$ & $3.78\pm0.30$ & $36.83\pm4.30$\\
			VS & $72.82\pm 0.22$ & $3.26\pm0.21$ & $18.30\pm1.31$\\
			TS & $71.53\pm0.13$ & $3.64\pm0.22$ & $32.74\pm2.17$\\
			CTS & $71.53\pm0.13$ & $3.21\pm0.26$ & $23.70\pm3.34$\\
			\bottomrule 
		\end{tabular}
		\caption{ Comparison of class-wise (VS, CTS) and non-class-wise (uncalibration, TS) calibration methods. There is 30\% label noise on classes 0-49. CTS and VS have lower ECE and max-ECE than the non-class-wise method, while VS changes the prediction accuracy.}
		\label{table:calibration}
	\end{center}\vs\vs
\end{table}

\vspace{3pt}
\noindent\textbf{\emph{The class-wise CTS and VS methods outperform non-class-wise methods.}} VS is another class-wise calibration method that may give better performance than non-class-wise methods, due to better fitting capability to each class (as long as overfitting does not occur).
In this set of simulations, we compare the class-wise CTS and VS methods with non-class-wise TS and uncalibrated methods.
We construct a training dataset with a 30\% label corruption rate for half of the classes (as described in Sec.~\ref{sec:setup}).
We compare the calibration error of VS, TS and CTS is according to accuracy, ECE, and max-ECE.
Table~\ref{table:calibration} shows the results.
In terms of max-ECE, VS is the most preferable, while CTS also has good performance.
In terms of ECE, CTS and VS outperform other methods due to their ability to handle heterogeneity via class-wise treatment. Table \ref{table:calibration10} shows the results on CIFAR-10 dataset with similar experimental setup. The improvements due to CTS and TS are noticeably more striking compared to CIFAR-100. We believe this is because CIFAR-10 is a relatively less complex dataset that is easier to (over)fit compared to CIFAR-100. Thus, calibration has a stronger impact.

\begin{figure}
	\centering
	\begin{subfigure}{0.4\textwidth}
		{\includegraphics[width=\textwidth]{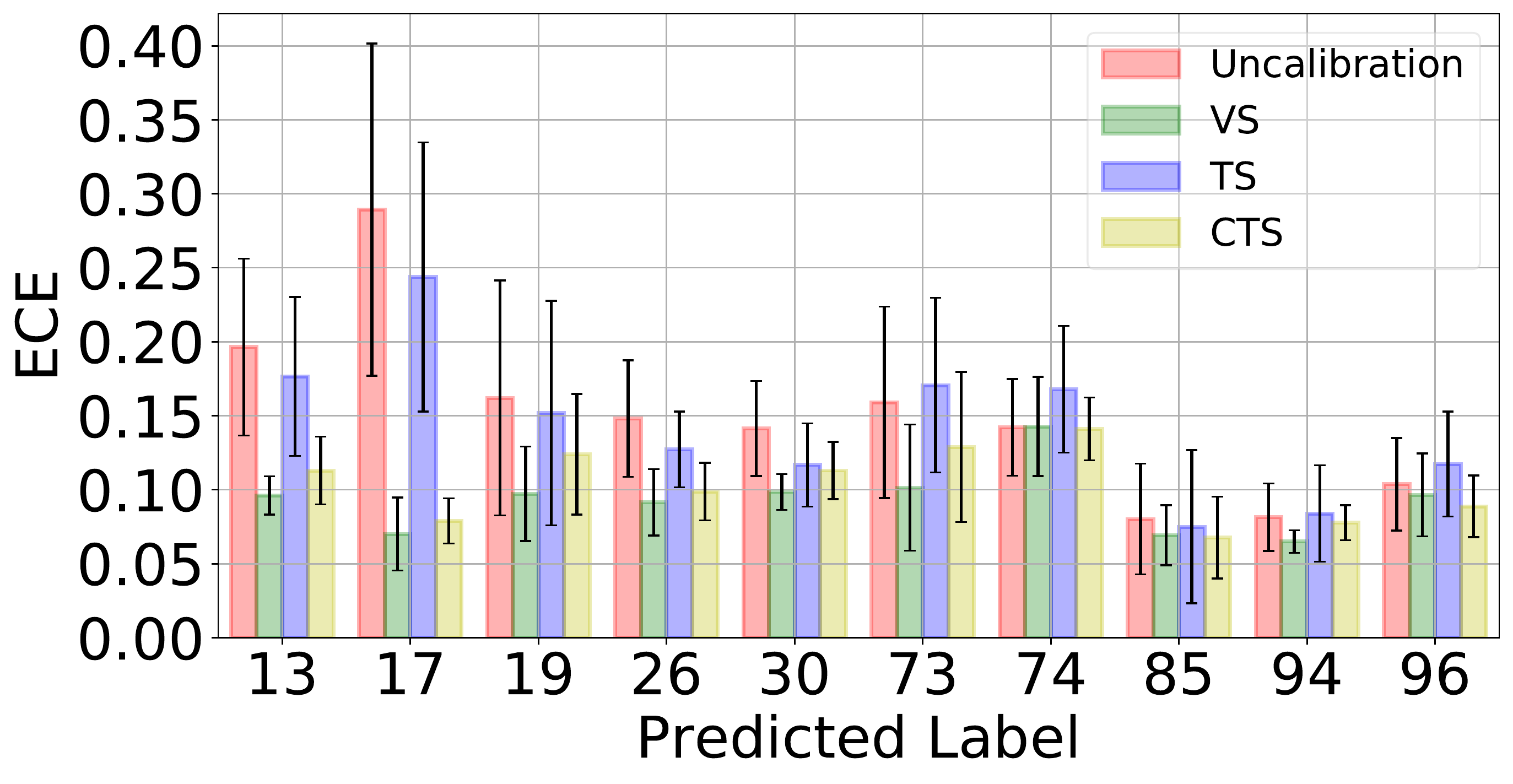}}\vspace{-5pt}
		\caption{ECE}
		\label{fig:per-class-ece}       
	\end{subfigure}
	\begin{subfigure}{0.4\textwidth}
		{\includegraphics[width=\textwidth]{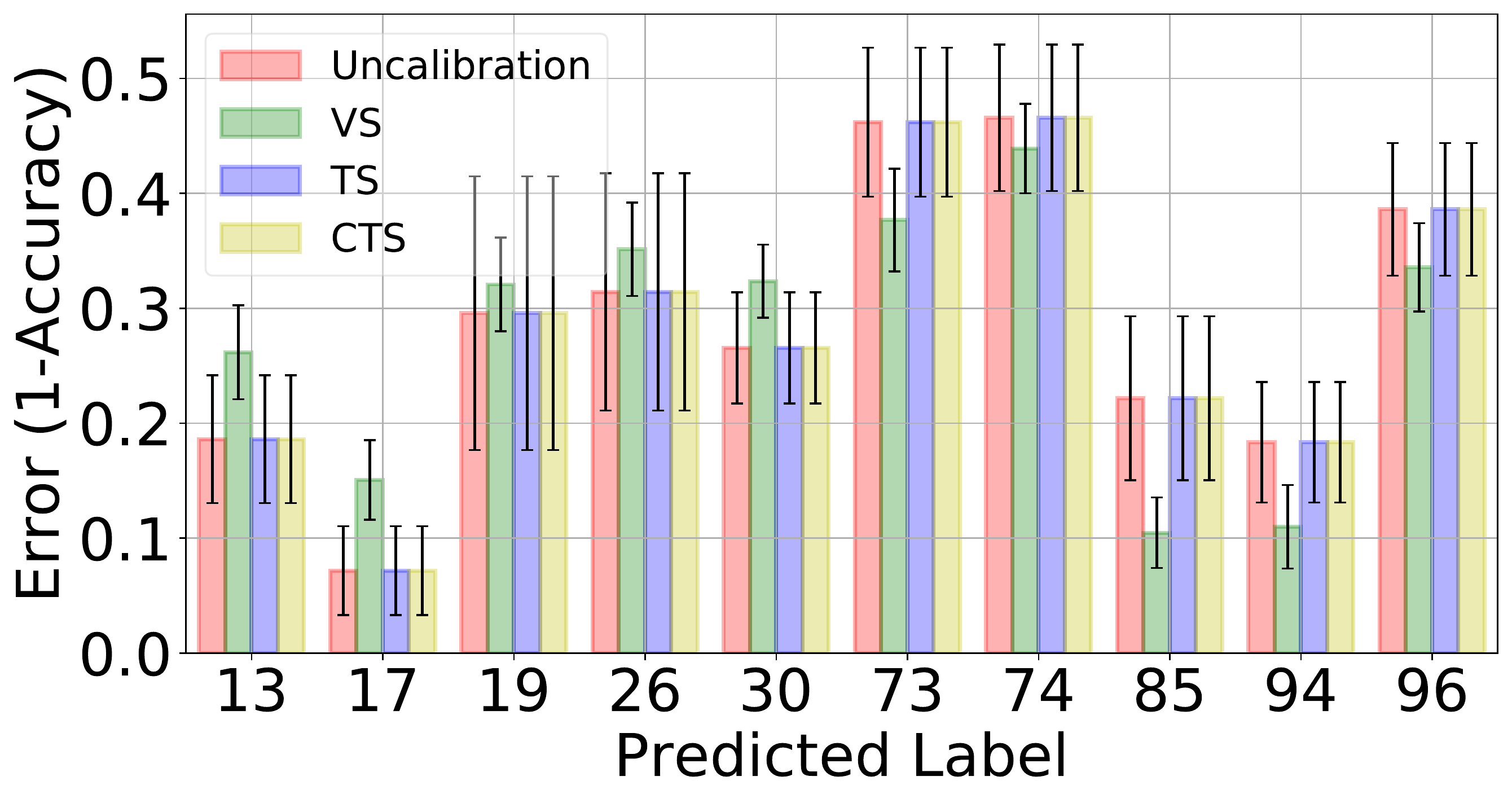}}\vspace{-5pt}
		\caption{Classification error}
		\label{fig:per-class-accuracy}
	\end{subfigure}
	\caption{ECE and accuracy for five random classes (from each of 0-49 and 50-99) are visualized. 
	(a) The class-wise algorithms (CTS, VS) have lower ECE than the traditional methods (uncalibrated, TS) in every class.
	(b) VS changes the classification accuracy, which may be undesirable.
	 }
	 \label{fig:per-class} \vs\vs
\end{figure}

\vspace{3pt}
\noindent\textbf{\emph{VS changes the predictions (in a good way!).}} Aside from the similar calibration performance of VS and CTS, VS {\em{improves the prediction accuracy}} by 1.29\%, which is a surprising observation.
However, modifying the predictions of the original classifier may be undesirable in certain applications, such as those with fairness concerns. For instance, as shown in Fig.~\ref{fig:per-class-accuracy}, VS uniformly degrades the prediction accuracy over noisy classes (classes 0-49) and uniformly improves the average accuracy over clean classes (classes 50-99). Note that noisy classes are already suffering from lower accuracy due to the noise, and VS ends up amplifying this while improving the overall accuracy. 
In contrast, by construction the CTS prediction is guaranteed to be consistent with the original classifier as discussed in Sec.~\ref{sec:approach}. We leave further investigation of VS and other calibration techniques that modify the classifier decision as an avenue for future investigation. Fig.~\ref{fig:per-class-ece} breaks down the results from Table~\ref{table:calibration}, and shows that ECE is lower for VS and CTS in every class when compared to TS and no calibration.


\begin{table}
	\begin{center}
		\begin{tabular}{l|c|c|r}
			\toprule 
			\textbf{Alg.} & \textbf{Acc. (\%)} & \textbf{ECE (\%)} & \textbf{max-ECE (\%)}\\
			\midrule 
			Uncal. &  $86.98 \pm 0.45$ & $6.50 \pm 0.46$ & $21.09 \pm 0.49$\\
			VS & $87.27 \pm 0.39$ & $1.31 \pm 0.11$ & $4.37 \pm 0.56$\\
			TS & $86.98 \pm 0.45$ & $3.69\pm 0.33$ & $13.85 \pm 0.64$\\
			CTS & $86.98 \pm 0.45$ & $1.19 \pm 0.25$ & $5.46 \pm 0.98$\\
			\bottomrule 
		\end{tabular}
		\caption{Experiments on CIFAR-10 trained with ResNet-20. 5 classes have 30\% label noise and 5 classes are clean. CIFAR-10 experiments show a more substantial benefit of calibration compared to CIFAR-100.}
		\label{table:calibration10}
	\end{center}\vs\vs
\end{table}

\subsection{Size Imbalanced Training Data}\vspace{0pt}
\label{sec:size}

\begin{figure}
	\centering
		{\includegraphics[width=0.42\textwidth]{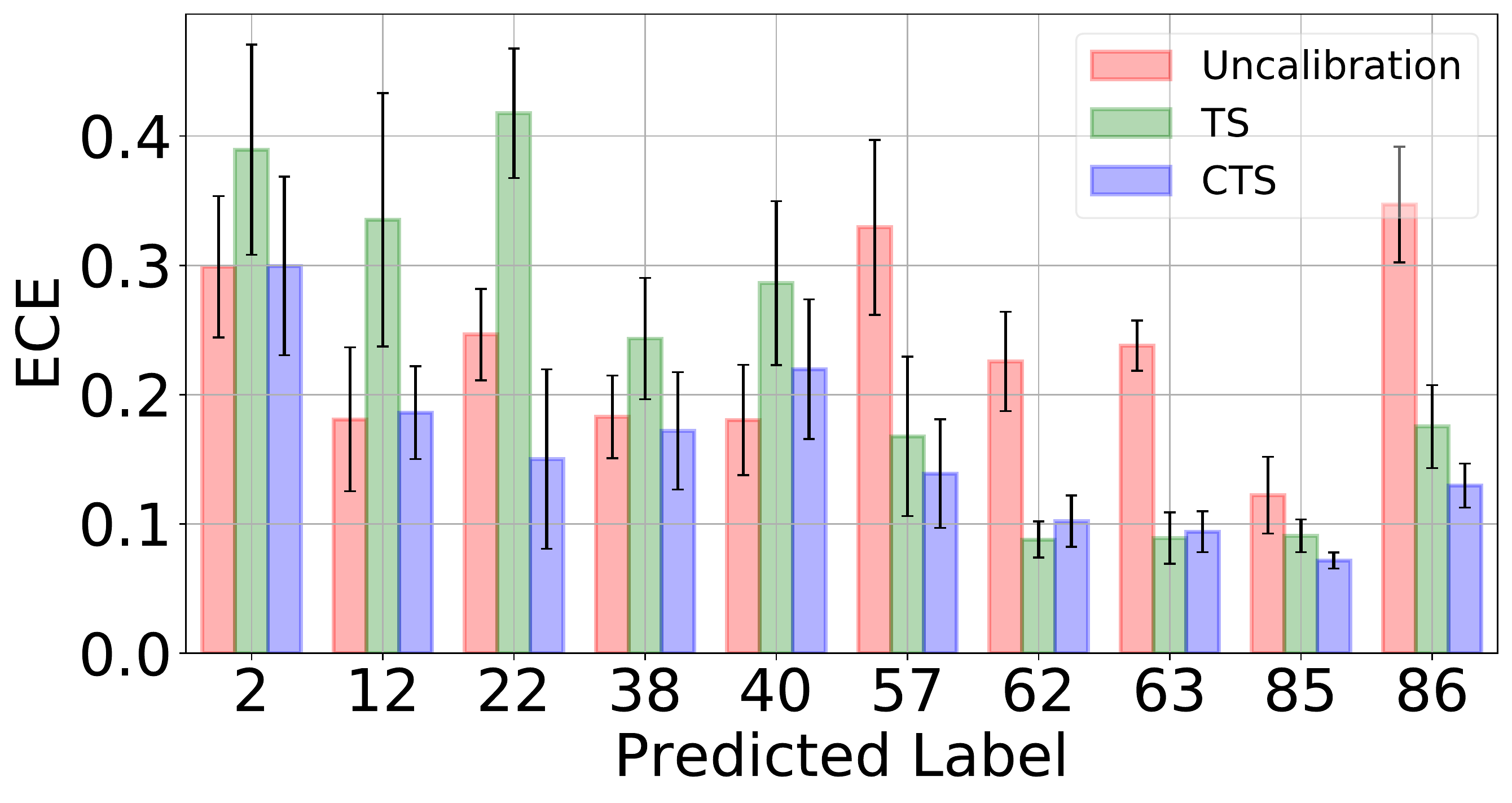}}
	\vspace{-0pt}
	\caption{\small{Smaller classes are sampled at 5\%. Five random classes from each of 0-49 and 50-99 are visualized. TS favors the large classes (at the expanse of small ones). CTS generally has lower ECE for both small and large classes. }}
	\label{fig balance}\vs\vs
\end{figure}\vspace{-0pt}

\textbf{\emph{Smaller training dataset results in over-confidence.}}
To verify the over-confidence on a small dataset, the complete standard training set (50k samples) of CIFAR-100 was undersampled without replacement, with sampling rates ranging from 5\%, 10\%, 60\% to 100\%. The reliability diagrams of the uncalibrated model are shown in Fig.~\ref{fig under samp} (finer resolution than Fig.~\ref{fig:conf_all}).
As the sampling rate decreases, the over-confidence of the model increases.
For lower sampling rates, the accuracy is lower (exact numbers are provided in the captions), showing that the over-confidence of the models trained on fewer samples is due to lower accuracy.


\vspace{3pt}
\noindent\textbf{\emph{CTS improves over the the uncalibrated classifier for imbalanced training data sizes.}}
We next investigate the effectiveness of CTS on the size unbalanced training set.
We construct the unbalanced training dataset as described in Sec.~\ref{sec:setup}, where smaller classes are $5\%$ as large as the non-down-sampled classes, containing only 27 training samples each. 
Fig.~\ref{fig balance} shows the ECE errors associated with individual classes as labeled by the classifier, \ie~$\ECE_k=\ECE(f,\Dc_k)$ where $\Dc_k$ is the conditional distribution $P(Y,X\bgl \hat{Y}=k)$. 
The results show that CTS provides uniform improvement over original uncalibrated classifier for all classes.
In contrast, TS actually inflates the calibration errors of the under-represented smaller classes, while improving the performance over larger classes.
This suggests that our proposed CTS provides a more fair treatment of the classes.

\begin{figure}
	\centering
		\begin{subfigure}{0.28\textwidth}	
		{\includegraphics[width=\linewidth]{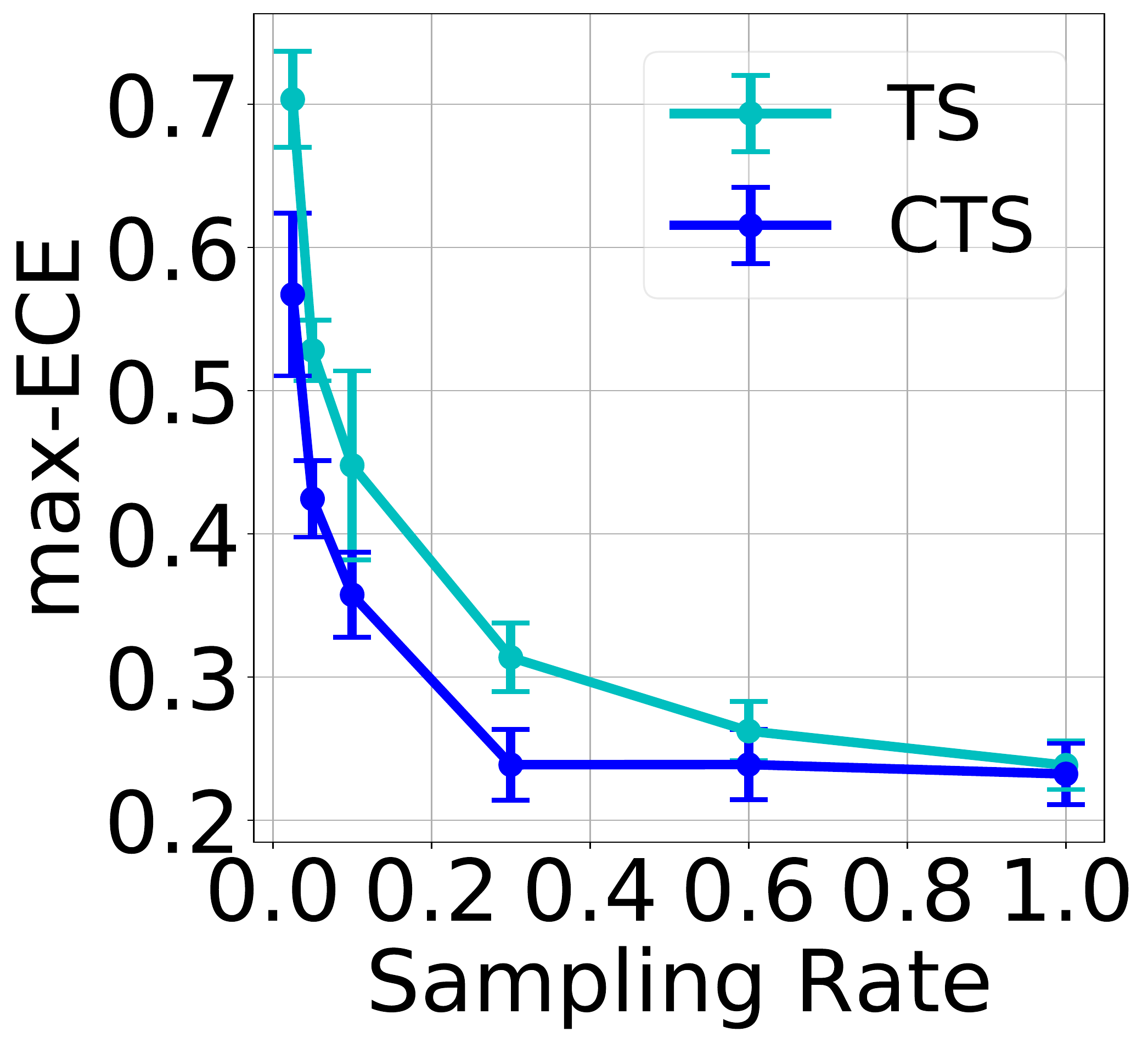}}
		\caption{max-ECE error}\label{fig unbalanceb}
	\end{subfigure}
	\begin{subfigure}{0.29\textwidth}
		{\includegraphics[width=\textwidth]{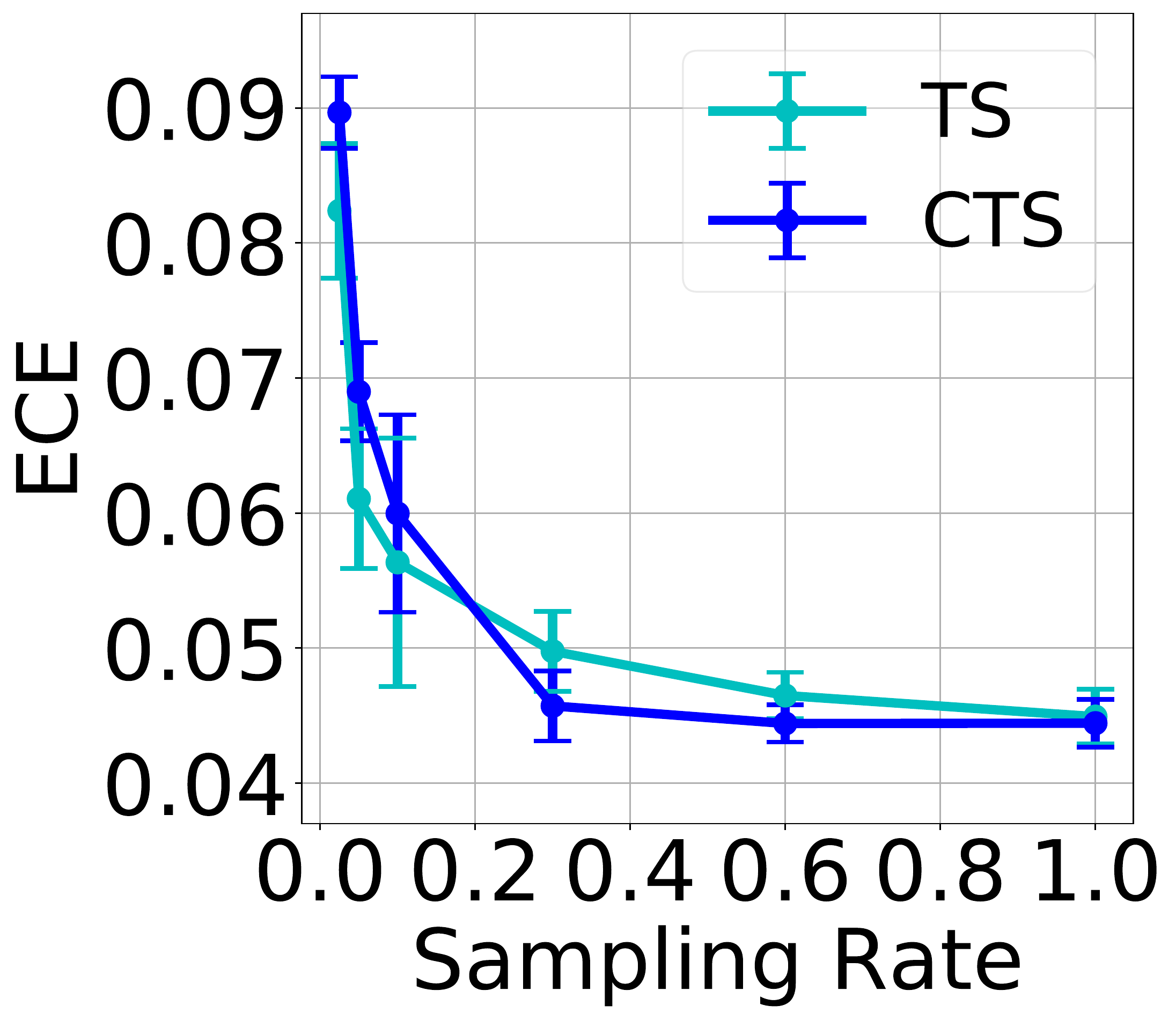}}    
		\caption{ECE error}\label{fig unbalancea}
	\end{subfigure}
	\begin{subfigure}{0.29\textwidth}	
		{\includegraphics[width = 1\linewidth]{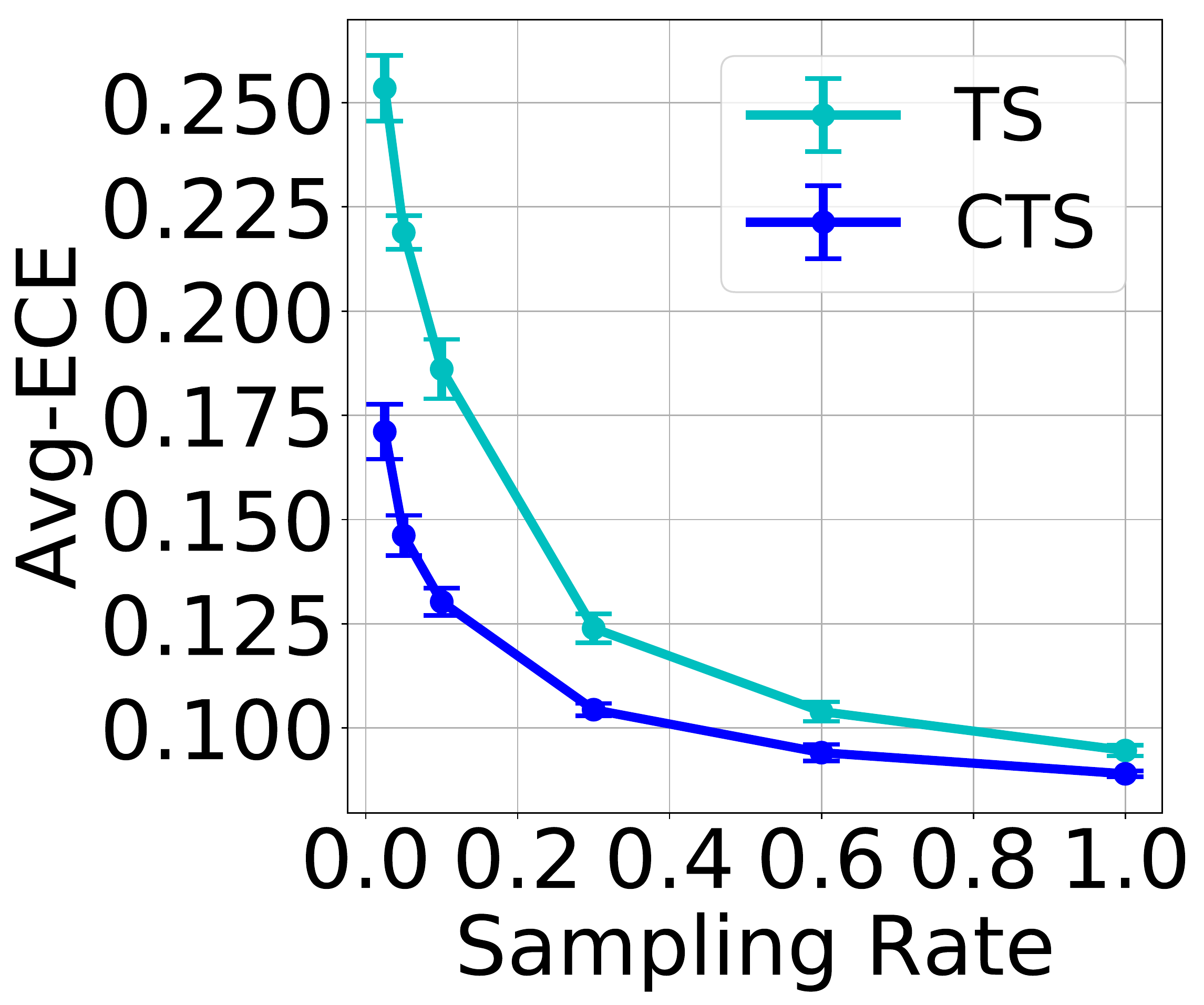}}
		\caption{Avg-ECE error}\label{fig unbalancec}
	\end{subfigure}	\vspace{-0pt}
	\caption{Calibration error as a function of the training set sampling rate.}\label{fig fig fig}\vspace{-0pt}
\end{figure}

\vspace{3pt}
\noindent\textbf{\emph{A closer look at the ECE metric.}} To further understand the impact of sample size on calibration error, we plot the ECE as a function of sampling rate in Fig.~\ref{fig fig fig}.
Fig.~\ref{fig unbalanceb} shows that CTS outperforms TS in terms of max-ECE metric for all sampling rates, highlighting the fairness benefit of CTS.
However, perhaps surprisingly, we find that in terms of overall ECE (where all samples are aggregated), TS sometimes outperforms CTS (Fig.~\ref{fig unbalancea}).
Upon digging deeper, we found that this is due to the way that individual class confidences output by TS combine in a favorable fashion when they are merged in a given confidence bin, as is done in the overall ECE metric.
For example, suppose there are two equal-sized classes, and fix a confidence bucket, \eg~$[0.4,0.6]$.
\begin{myitemize}\vspace{-0pt}
\item Suppose Class 1 has average accuracy of 0.52, TS confidence of 0.6, and CTS confidence of 0.54.
\item Suppose Class 2 has average accuracy of 0.48, TS confidence of 0.4, and CTS confidence of 0.5.\vspace{-0pt}
\end{myitemize}
In this case, TS will achieve $\ECE_1=\ECE_2=0.08$ whereas CTS will achieve $\ECE_1=\ECE_2=0.02$, so CTS seems better. However, CTS is overconfident in both classes whereas TS is perfectly calibrated when both classes are combined, resulting in aggregated $\ECE_{TS}=0$ and $\ECE_{CTS}=0.02$.

This indicates that the ECE metric is unable to capture how well each class is doing, and demonstrates the usefulness of worst-case metrics such as max-ECE. On the other hand, it also highlights that optimizing calibration over all samples (\ie~TS compared to CTS with $\Gamma=\infty$) can have favorable properties by aligning the over-/under-confidence directions. Motivated by this, we consider an alternative metric: namely, the average of the individual ECEs, as given by:\vs
\[
\text{Avg-ECE}=\frac{1}{K}\sum_{k=1}^K\ECE(f,\Dc_k).
\]
The comparison of the Avg-ECE's is shown in Fig.~\ref{fig unbalancec}. This figure shows a rather substantial performance difference, and thus highlights how CTS does a much better job than TS when focused on the performance across individual classes. 


\vspace{-0pt}\section{Conclusions}
\label{sec:conclusions}\vspace{-0pt}

In this paper, we investigated the impacts of noise, sample size, and heterogeneity of the training data on the confidence of the machine learning prediction outputs. We provided empirical evidence and theory showing that noisiness in training data leads to under-confidence and fewer samples leads to over-confidence. To address these class-wise imbalance issues, we proposed class-wise calibration techniques, which, when applied to the temperature-scaling method, leads to our proposed class-wise temperature scaling algorithm.
We provided validation sample complexity bounds for our proposed scheme, and our numerical simulations demonstrate the efficacy of these class-wise techniques.
Future work includes examining the effectiveness of class-wise calibration beyond temperature scaling and on more complex and wider range of datasets, as well as developing class-wise algorithms for jointly optimizing accuracy and calibration during training.



\bibliographystyle{plain}
\bibliography{icml}

\appendix

\section{CIFAR-10 Experiments}
The remaining experiments and figures are on CIFAR-10 dataset and closely follow the setup of CIFAR100.  In our experiments, whenever validation
is needed, the original training set is split into 45k
training samples and 5k validation samples. We only
modify the training data. The validation set is always clean (i.e., not noisy), and validation class sizes
are equal with 500 samples each. All experiments use
the standard data augmentation. Experiments are repeated five times
with different random seeds. To evaluate the impact
of heterogeneous data, we construct two variants of CIFAR-10:

\begin{itemize}
\item {\bf{Noise-imbalanced dataset construction:}} In the training dataset, we add label noise to classes
0 to 4 with noise rate $\rho$ varying from 0 to 1. Noise
rate $\rho$ means that with $\rho$ probability a label is randomly assigned to one of the 10 labels. Classes 5
to 9 remained unchanged. This results in a partially noisy training set.

\item {\bf{Size-imbalanced dataset construction:}} We
under-sample the classes 0 to 4 in the training set,
with the sampling rate $\rho$ varying from 0.01 to 1.
Instead of the usual 4500 training samples, undersampled classes have only $4500\rho$ training samples.
For instance, with $\rho = 0.01$, the smaller classes
contain only 45 samples resulting in a highly unbalanced dataset. The overall training set is obtained by combining the 5 downsampled classes
and the other 5 classes. The validation set remain unchanged (same size for every class).
\end{itemize}

\noindent{\bf{Neural network model:}} To perform image classification, we utilize the ResNet-20 network model.

\noindent {\bf{Training:}} 200 training epochs of the ResNet-20 with
Adam optimizer were used to fit the data, with cross-entropy as the loss function using Keras and TensorFlow (Chollet et al. (2015); Abadi et al. (2016)). The initial learning rate is initially set to $10^{-3}$ and decreases to $10^{-4}$ after 80 epochs.

\begin{figure*}
\centering
		\begin{subfigure}[t]{0.27\textwidth}
			\includegraphics[width=\linewidth]{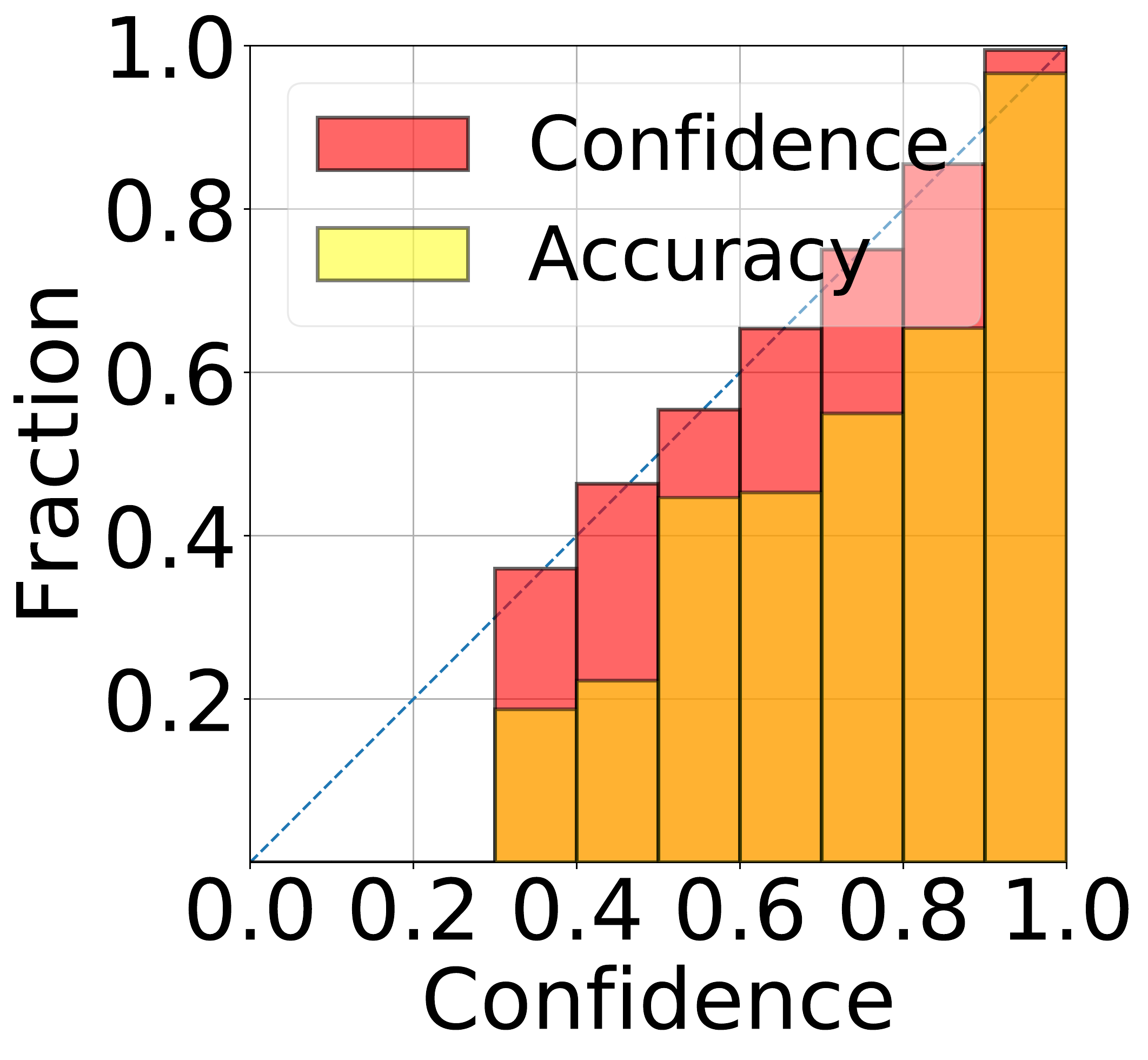}
			\caption{Standard CIFAR-10 model trained with clean data}\label{fig1a10}
		\end{subfigure}
		~
		\begin{subfigure}[t]{0.27\textwidth}
			\includegraphics[width=\linewidth]{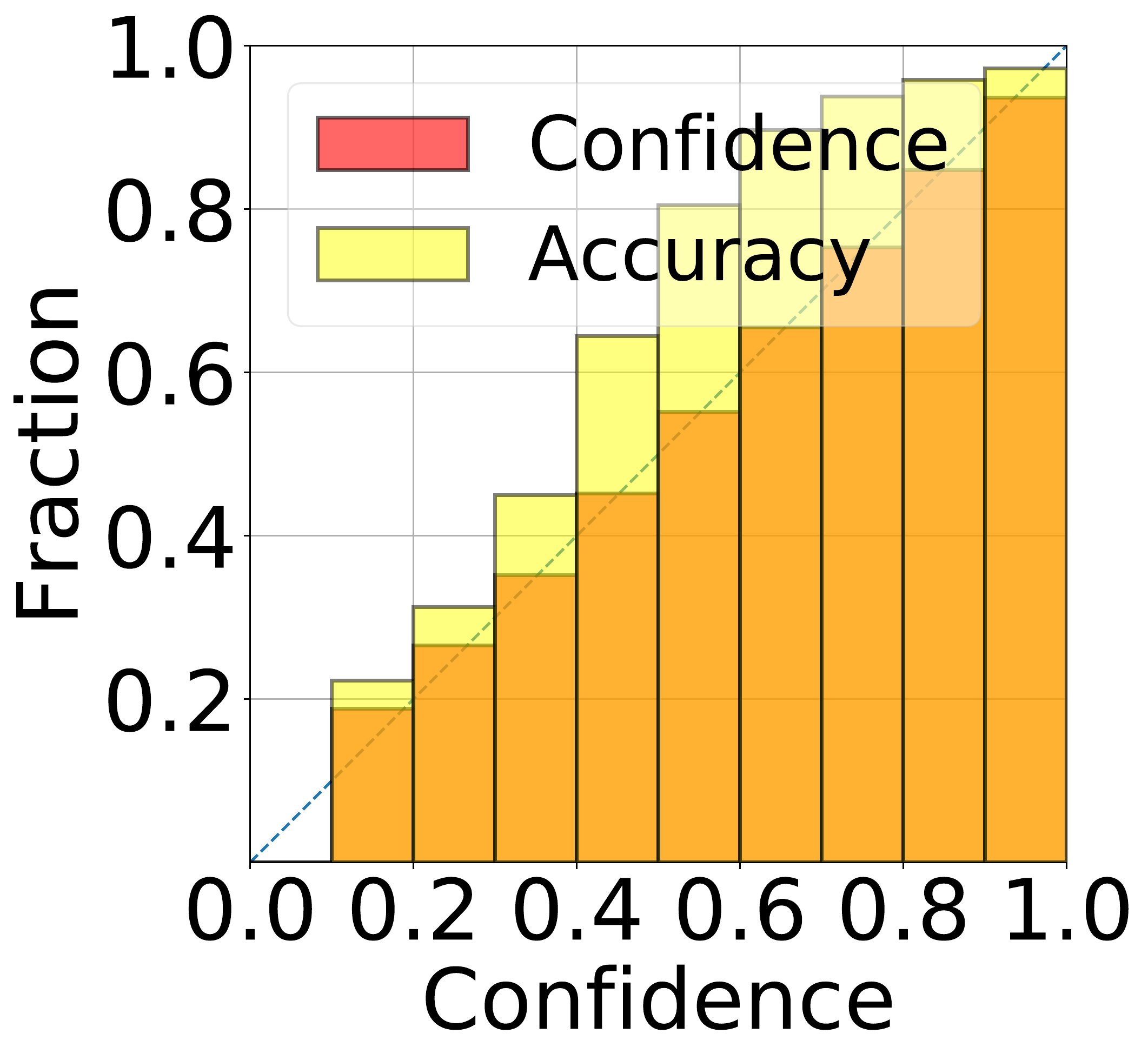}
			\caption{CIFAR-10 model trained with noisy data (30\% chance of label corruption)}\label{fig1b10}
		\end{subfigure}
		~
				\begin{subfigure}[t]{0.27\textwidth}
			\includegraphics[width=\linewidth]{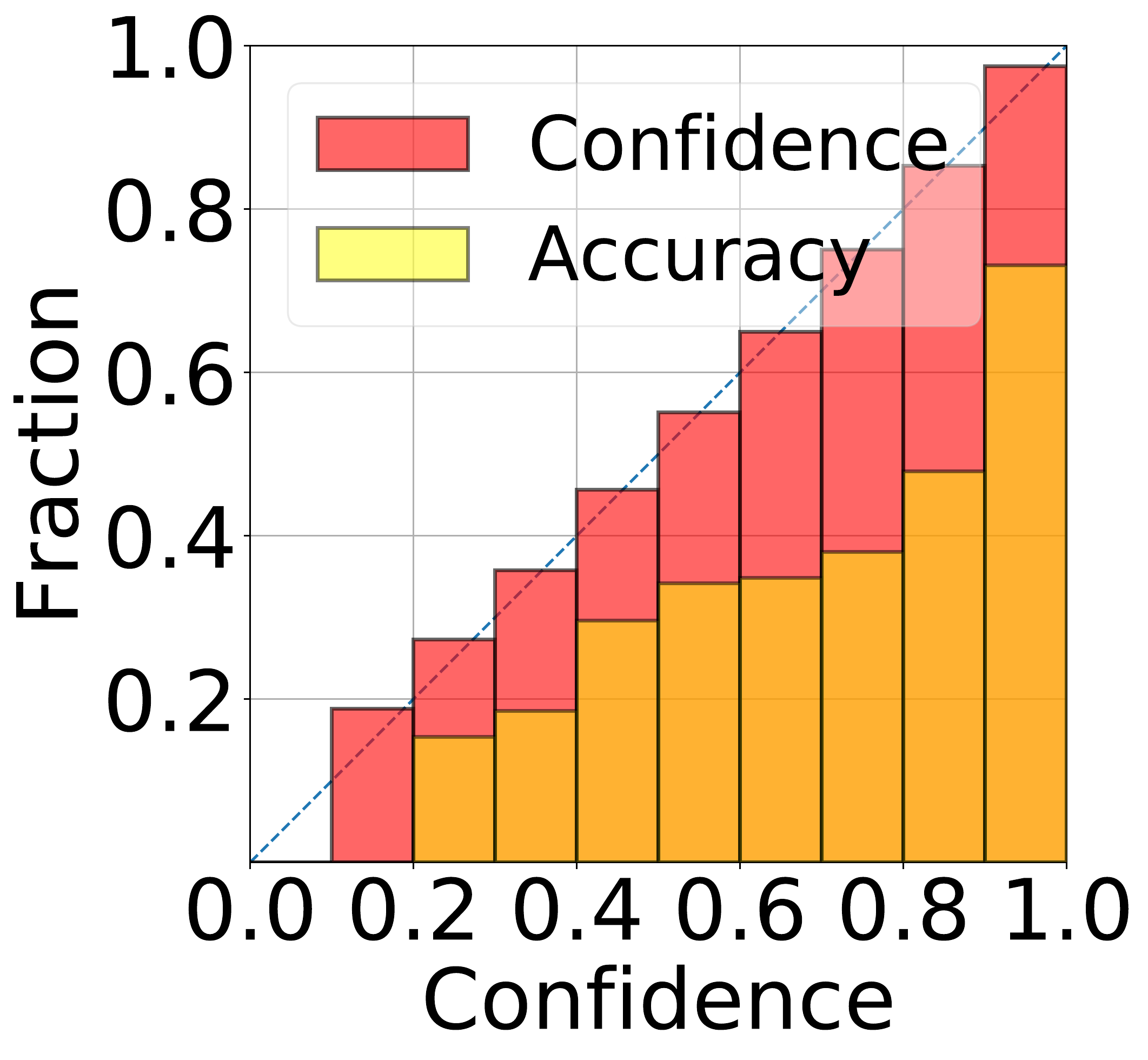}
			\caption{CIFAR-10 model trained with small classes (100 labels per class rather than 5000)}\label{fig1c10}
		\end{subfigure}\vspace{-3pt}
	\caption{ 
	Reliability diagrams for CIFAR-10, generated by binning the test results by confidence values, and plotting the confidence (red), accuracy (yellow), and their overlap (orange). 
	With perfect calibration, the accuracy should align exactly with the confidence along the diagonal. 
	The model (a) trained with clean data is over-confident (higher confidence than accuracy); (b) trained with noisy data is under-confident (lower confidence than accuracy); and (c) trained with small data size is more over-confident than (a).
	}
	\label{fig:conf_all10}\vspace{-5pt}
\end{figure*}

\begin{figure}
	\centering
	\begin{subfigure}{0.35\textwidth}
			{\includegraphics[width=\textwidth]{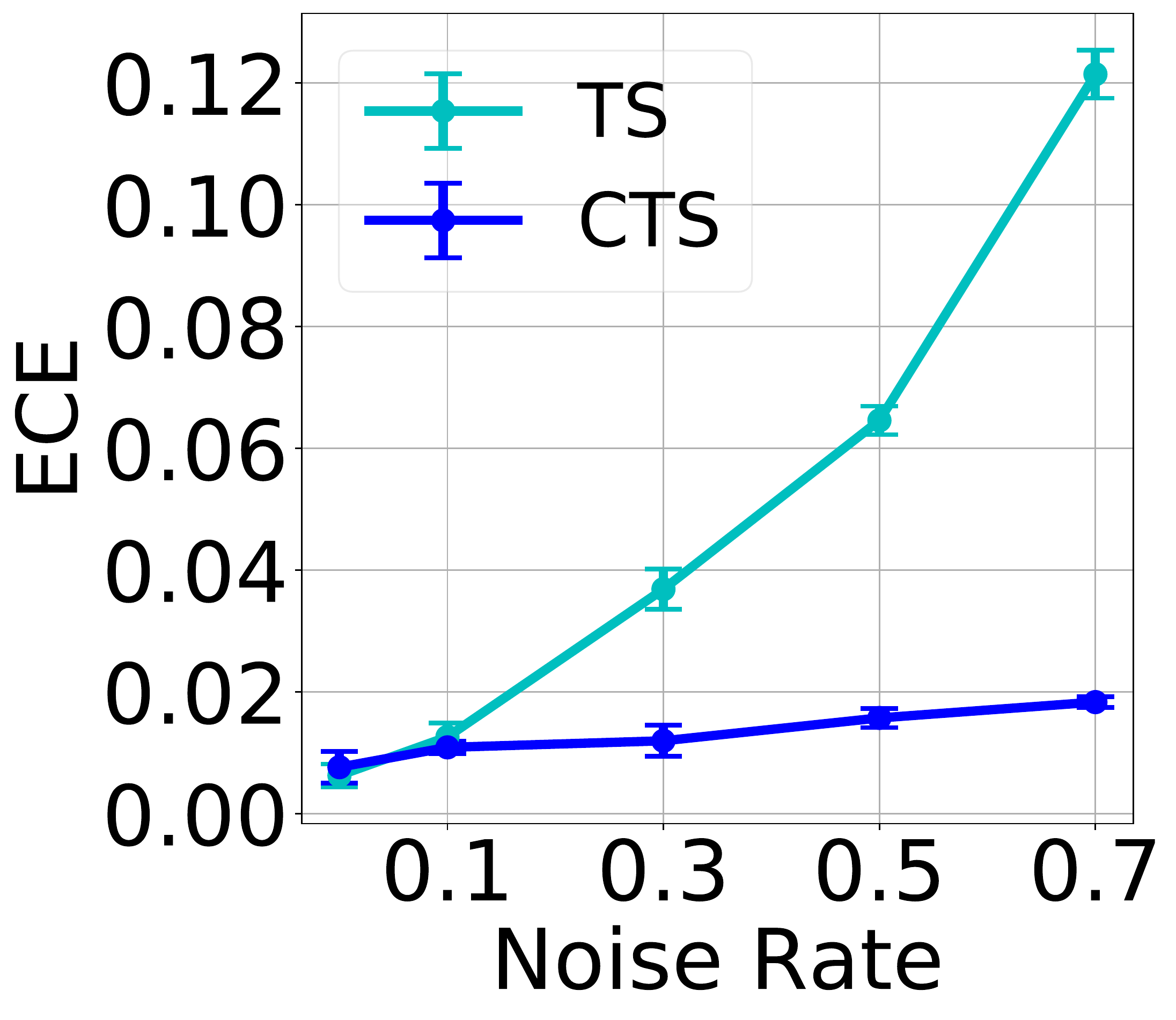}}
			\caption{ECE}       
	\end{subfigure}
	\begin{subfigure}{0.33\textwidth}
			{\includegraphics[width=\textwidth]{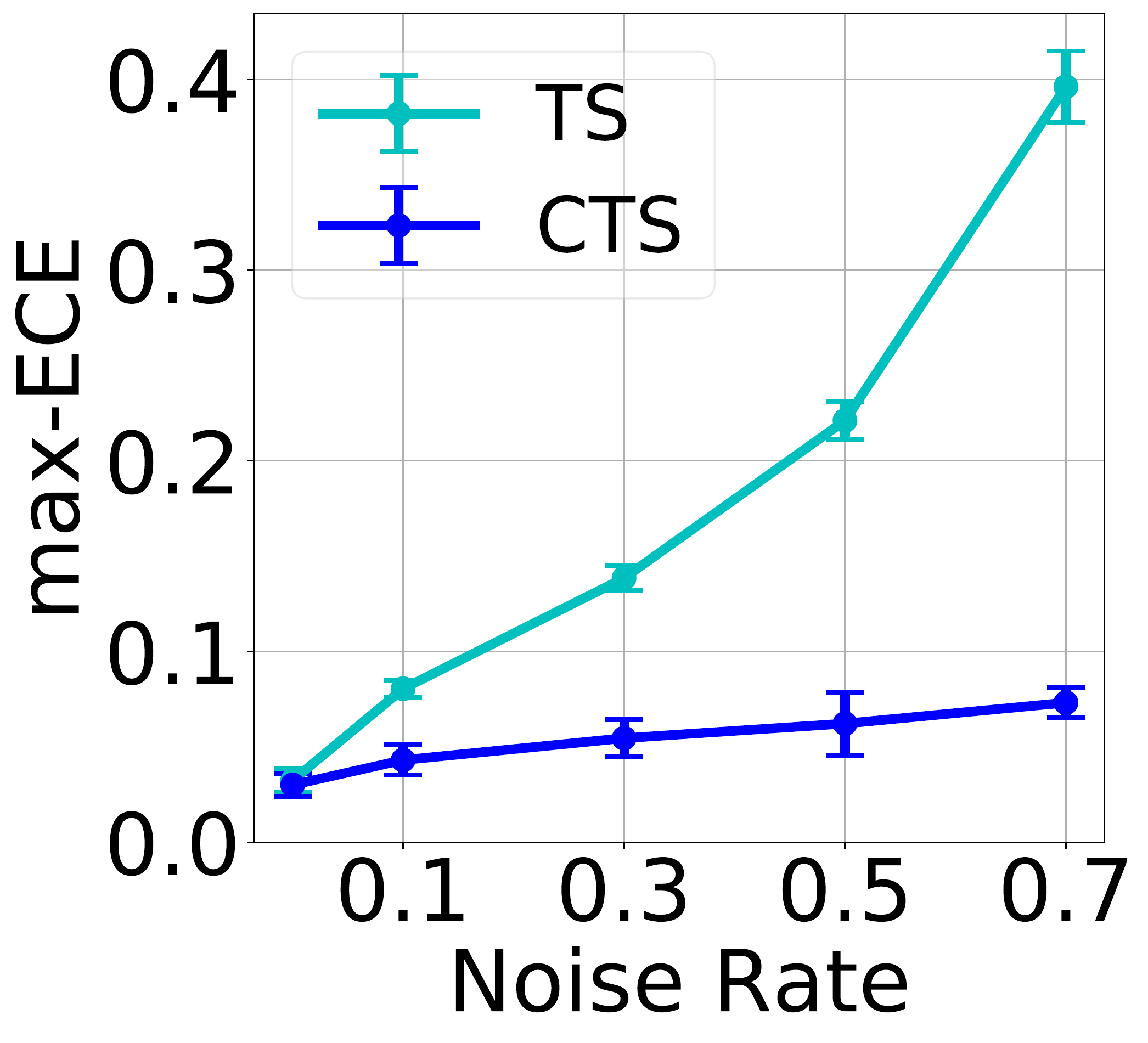}}
			\caption{max-ECE}       
	\end{subfigure}
	\caption{Noise imbalance: Impact of training data noise on the TS and CTS methods. CTS has lower error in terms of both the ECE and max-ECE metrics.}
	\label{fig:noise10}
	
\end{figure}

\begin{figure}
	\centering
	\begin{subfigure}{0.45\textwidth}
		{\includegraphics[width=\textwidth]{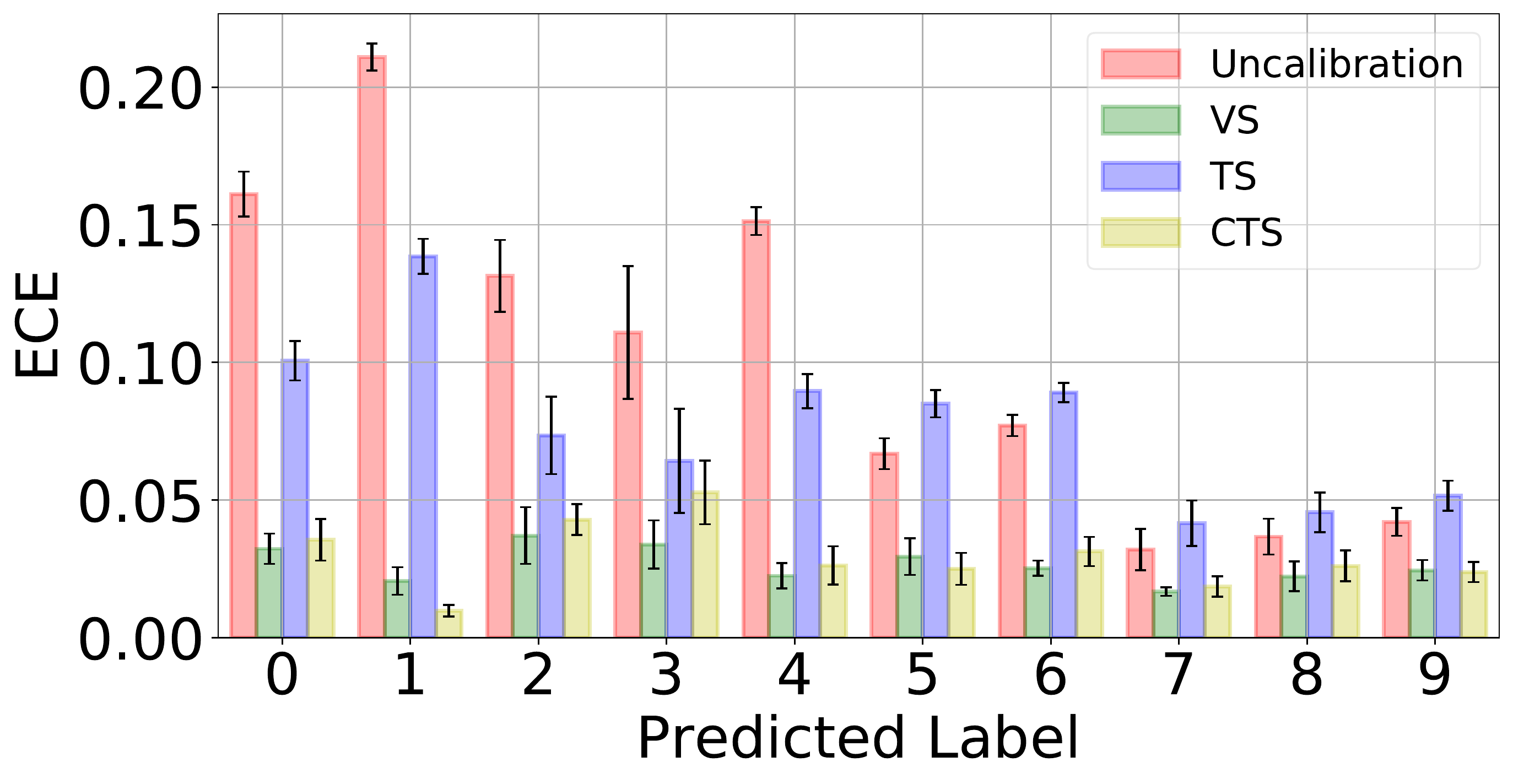}}\vspace{-5pt}
		\caption{ECE}
		\label{fig:per-class-ece}       
	\end{subfigure}
	\begin{subfigure}{0.45\textwidth}
		{\includegraphics[width=\textwidth]{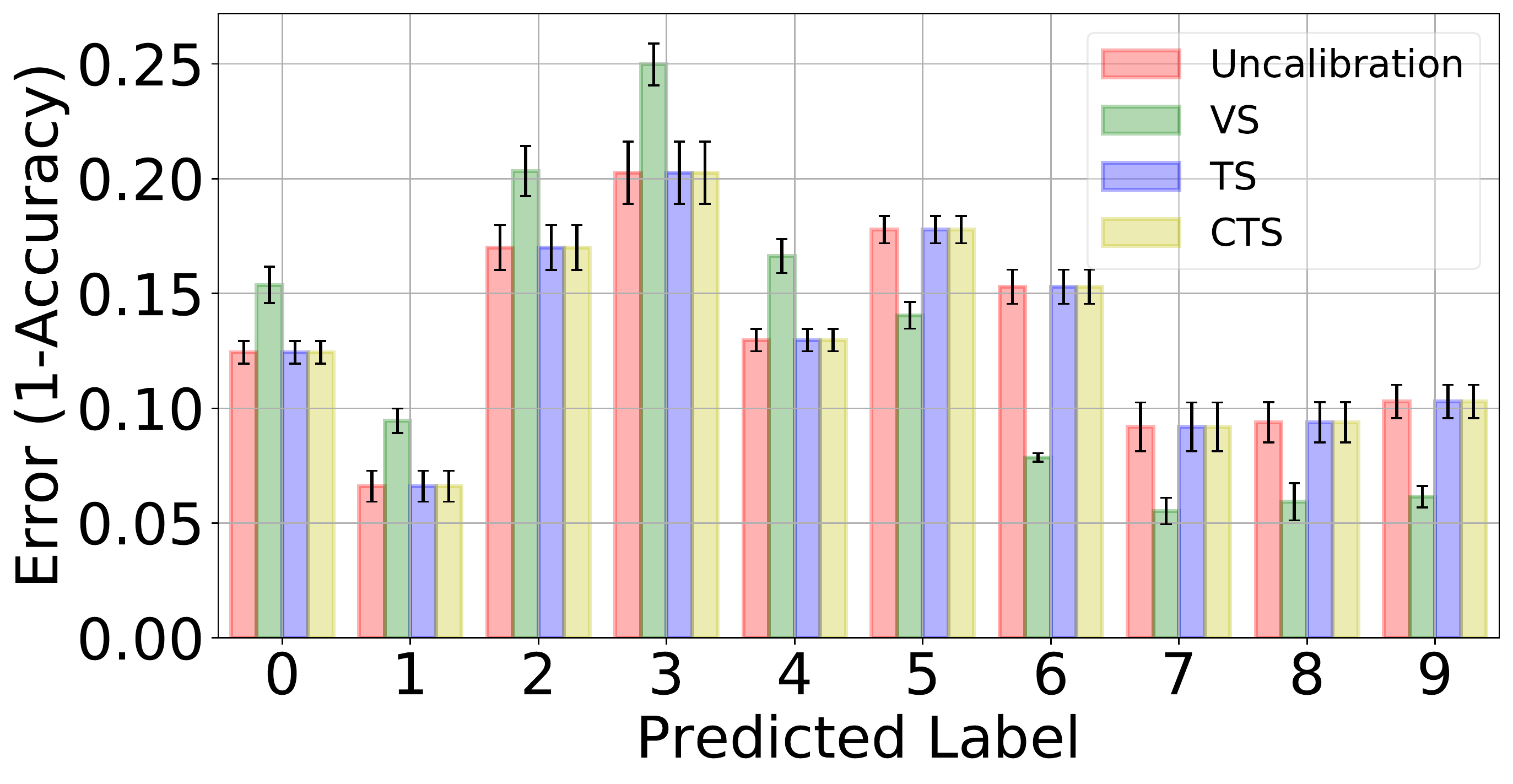}}\vspace{-5pt}
		\caption{Classification error}
		\label{fig:per-class-accuracy}
	\end{subfigure}
	\caption{Noise imbalance: Per-class error in terms of ECE and classification accuracy.
	(a) The class-wise algorithms (CTS, VS) have lower ECE than the traditional methods (uncalibrated, TS) in every class.
	(b) VS changes the classification accuracy, which may be undesirable.
	 }
	 \label{fig:per-class10} \vs\vs
\end{figure}

\begin{figure}
 	\centering
 	\begin{subfigure}{0.23\textwidth}
 			\includegraphics[width=\textwidth]{figs_CIFAR10/sample_intro/sample_fraction002}
 			\caption{2\% sampling rate, 80.5\% confidence, 53.8\% accuracy}
 	\end{subfigure}
 	~
 	\begin{subfigure}{0.23\textwidth}
 			\includegraphics[width=\textwidth]{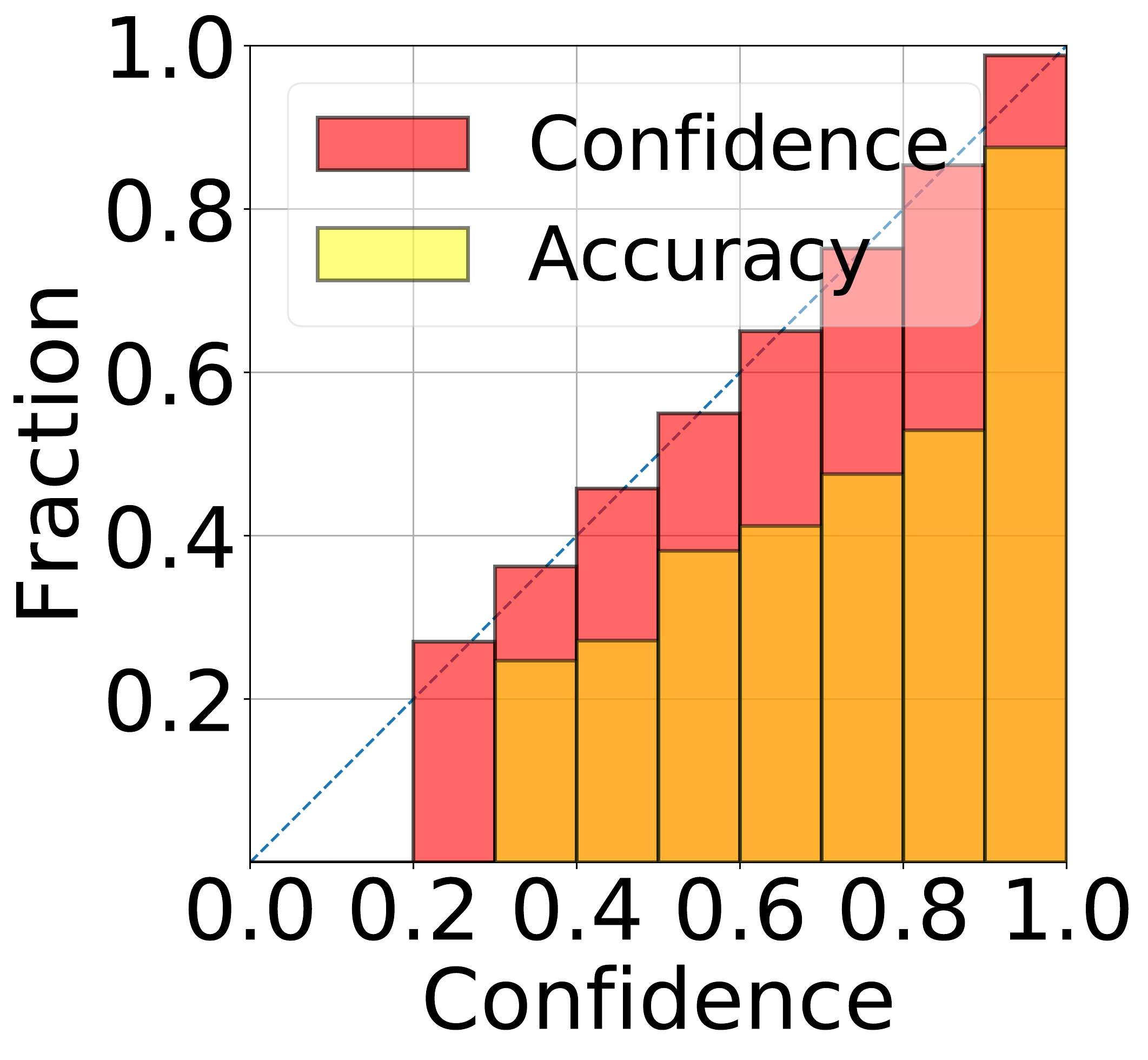}
 			\caption{10\% sampling rate, 90.5\% confidence, 75.7\% accuracy}
 	\end{subfigure}
  	\begin{subfigure}{0.23\textwidth}
 		\includegraphics[width=\textwidth]{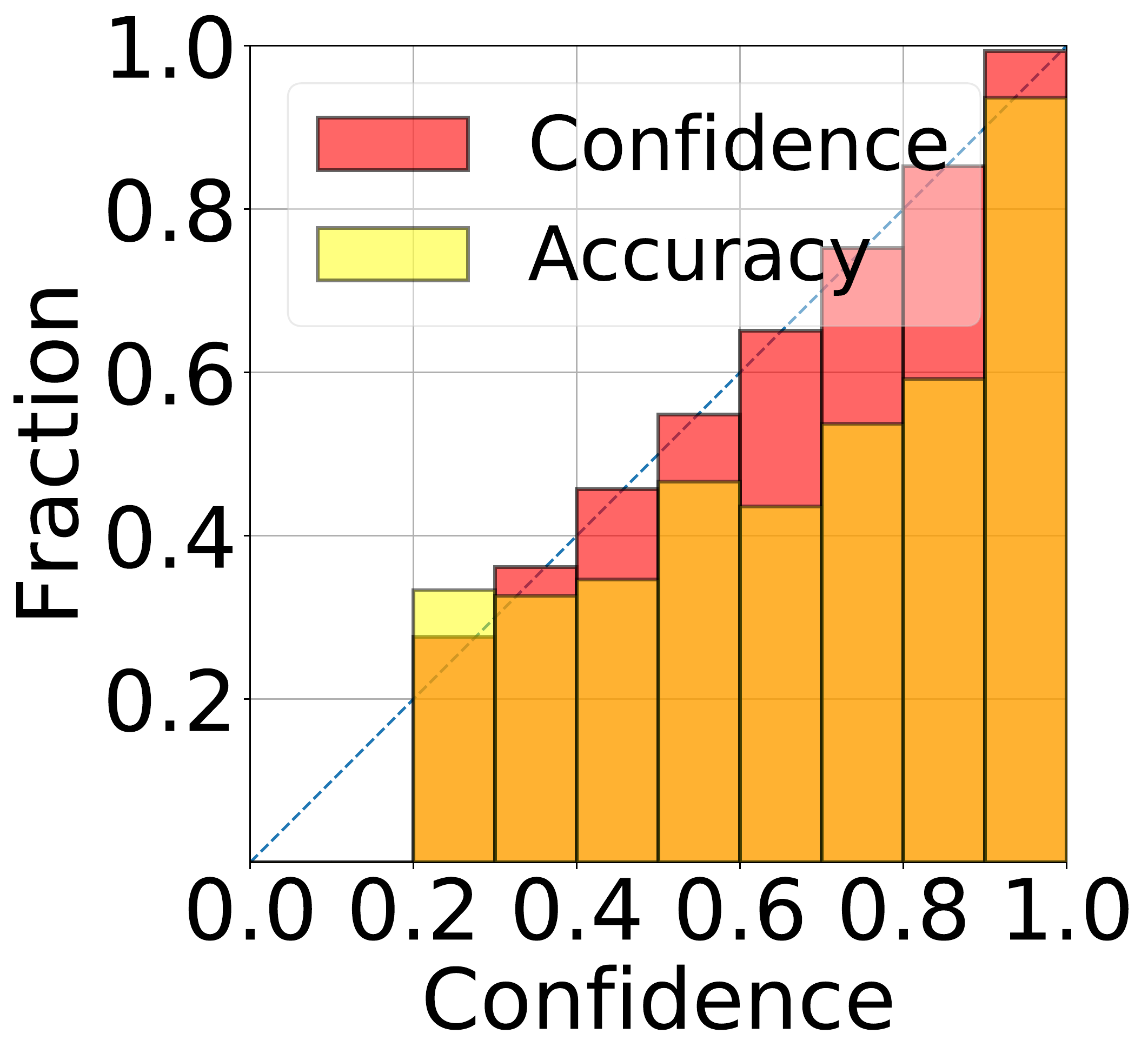}
 		\caption{40\% sampling rate, 94.7\% confidence, 86.9\% accuracy}
 	\end{subfigure}
 	~
 	\begin{subfigure}{0.23\textwidth}
 		\includegraphics[width=\textwidth]{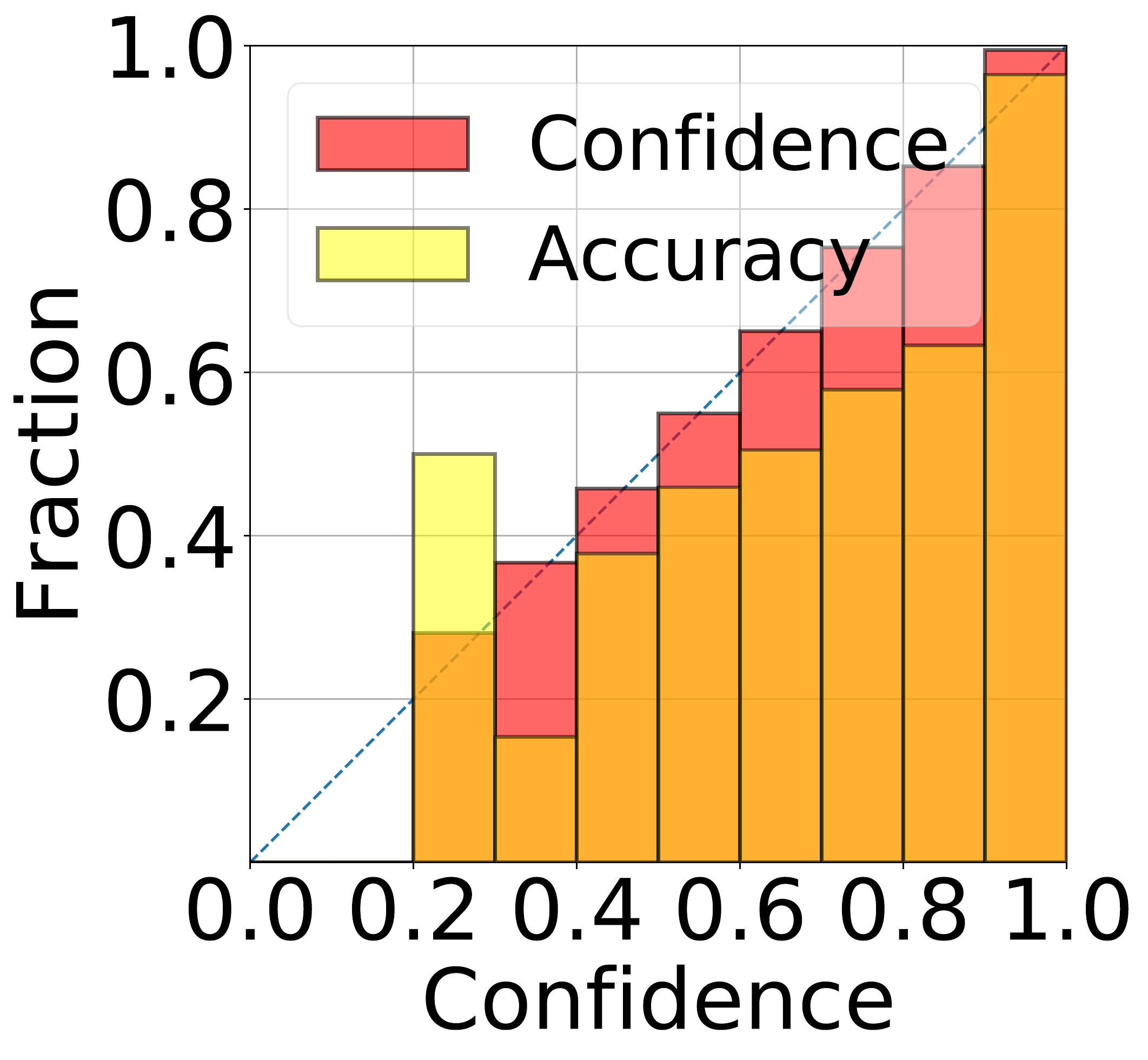}
 		\caption{100\% sampling rate, 96.0\% confidence, 91.5\% accuracy}
 	\end{subfigure}
 	\caption{Sample imbalance: Reliability diagram and over-confidence for different training dataset sizes for CIFAR-10. A smaller training dataset size leads to further over-confidence. This is because the model accuracy degrades rapidly with fewer samples. }\label{fig under samp10}
 \end{figure}

The outcomes of the experiments parallel the outcomes of CIFAR-100 experiments. However the calibration has a more dramatic impact on CIFAR-10 as discussed previously (e.g.~contrasting Tables \ref{table:calibration} and \ref{table:calibration10}).  Figure \ref{fig:conf_all10} is generated in a similar fashion to and is counterpart of Figure \ref{fig:conf_all}. CIFAR-10 experiments on noisy dataset are illustrated by Figures \ref{fig:noise10} and \ref{fig:per-class10}. These figures are created in a similar fashion to Figures \ref{fig:noise} and \ref{fig:per-class}. CIFAR-10 experiments on size-unbalanced dataset are illustrated by Figures \ref{fig under samp10} and \ref{fig fig fig10}. Figure \ref{fig under samp10} is generated in a similar fashion to Figure \ref{fig under samp} for different level of under-samplings (size-imbalancedness level). Finally, Figure \ref{fig fig fig10} is created in a similar way to Figure \ref{fig fig fig}.

\vspace{-15pt}




\begin{figure}
	\centering
		{\includegraphics[width=0.65\textwidth]{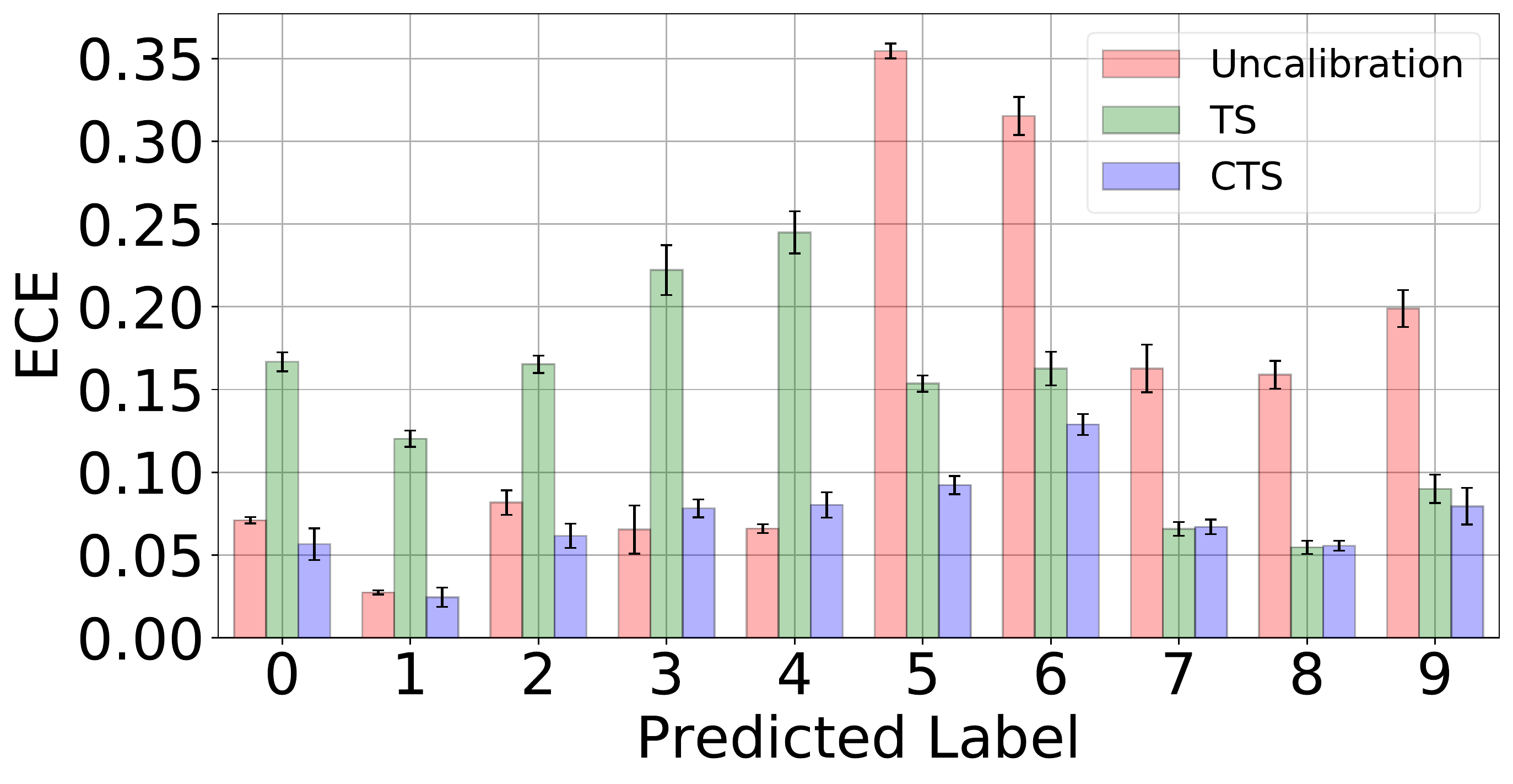}}
	\caption{Sample imbalance: Per-class ECE when the first five classes are
downsampled at 6\%. TS heavily favors the large classes (at the expanse of small ones). CTS generally has lower ECE
for both the small and large classes.}
	\label{fig balance10}\vs\vs
\end{figure}

\begin{figure}
	\centering
		\begin{subfigure}{0.32\textwidth}	
		{\includegraphics[width=\linewidth]{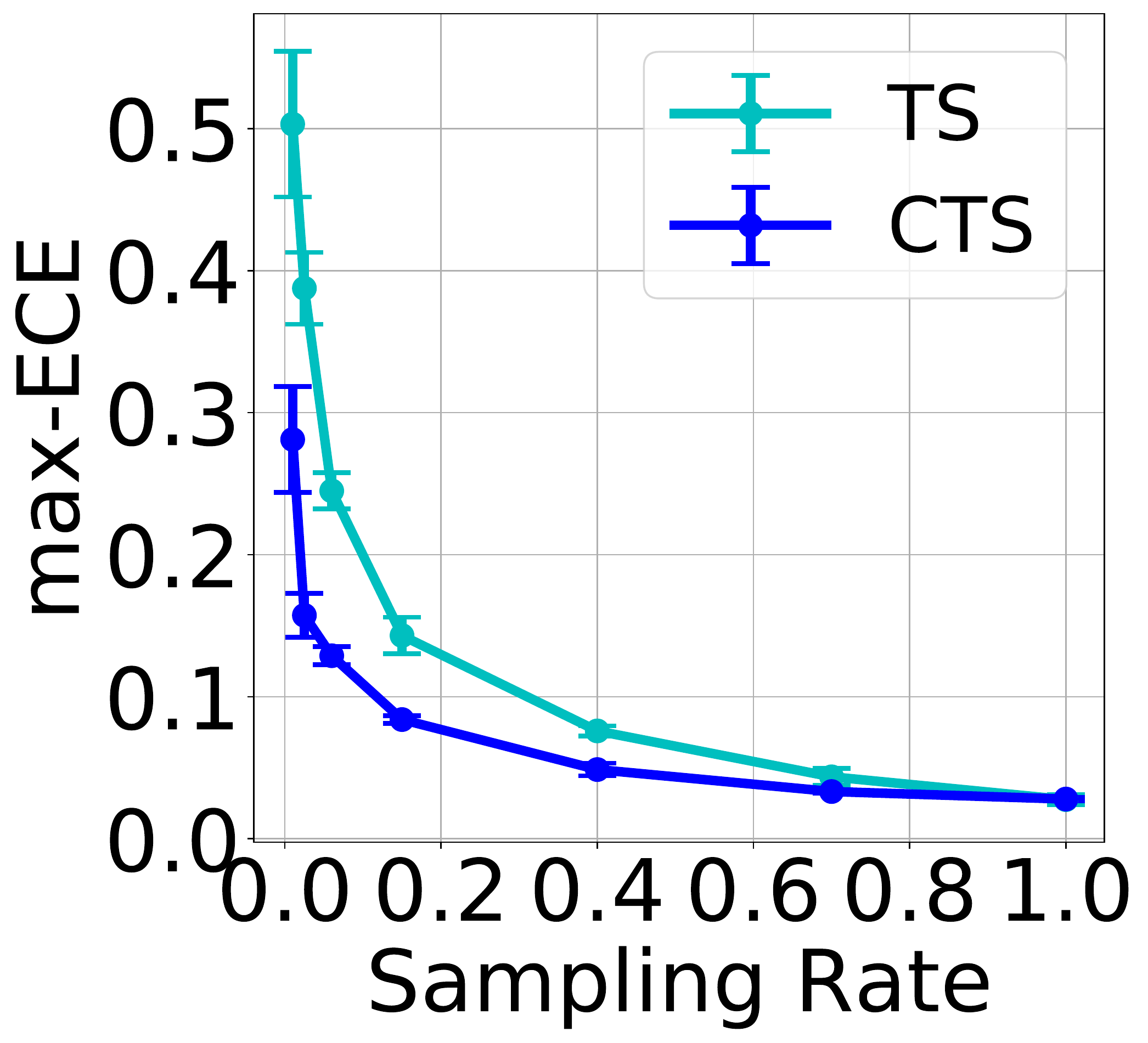}}
		\caption{max-ECE error}\label{fig unbalanceb}
	\end{subfigure}
	\begin{subfigure}{0.33\textwidth}
		{\includegraphics[width=\textwidth]{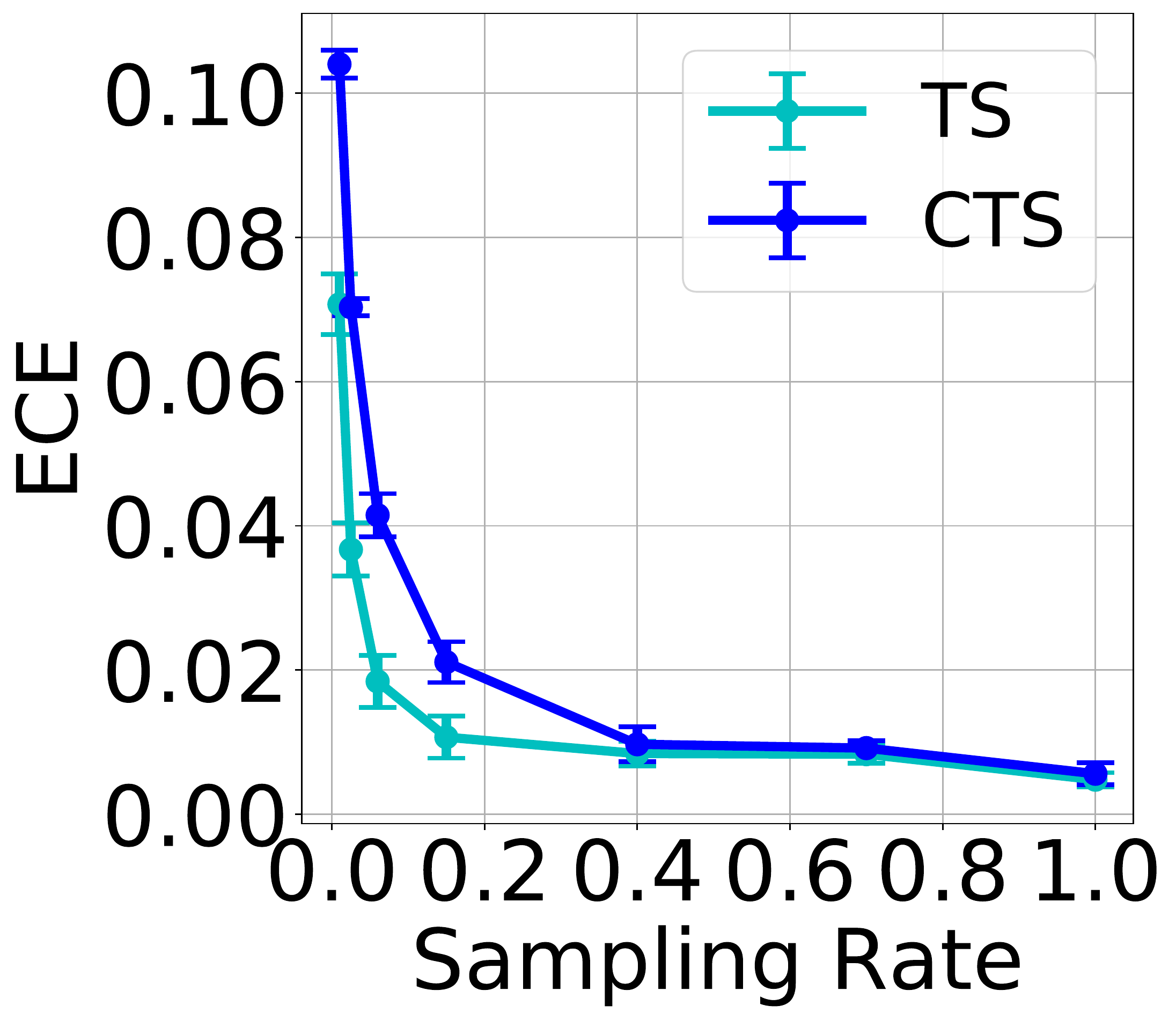}}    
		\caption{ECE error}\label{fig unbalancea}
	\end{subfigure}
	\begin{subfigure}{0.33\textwidth}	
		{\includegraphics[width = 1\linewidth]{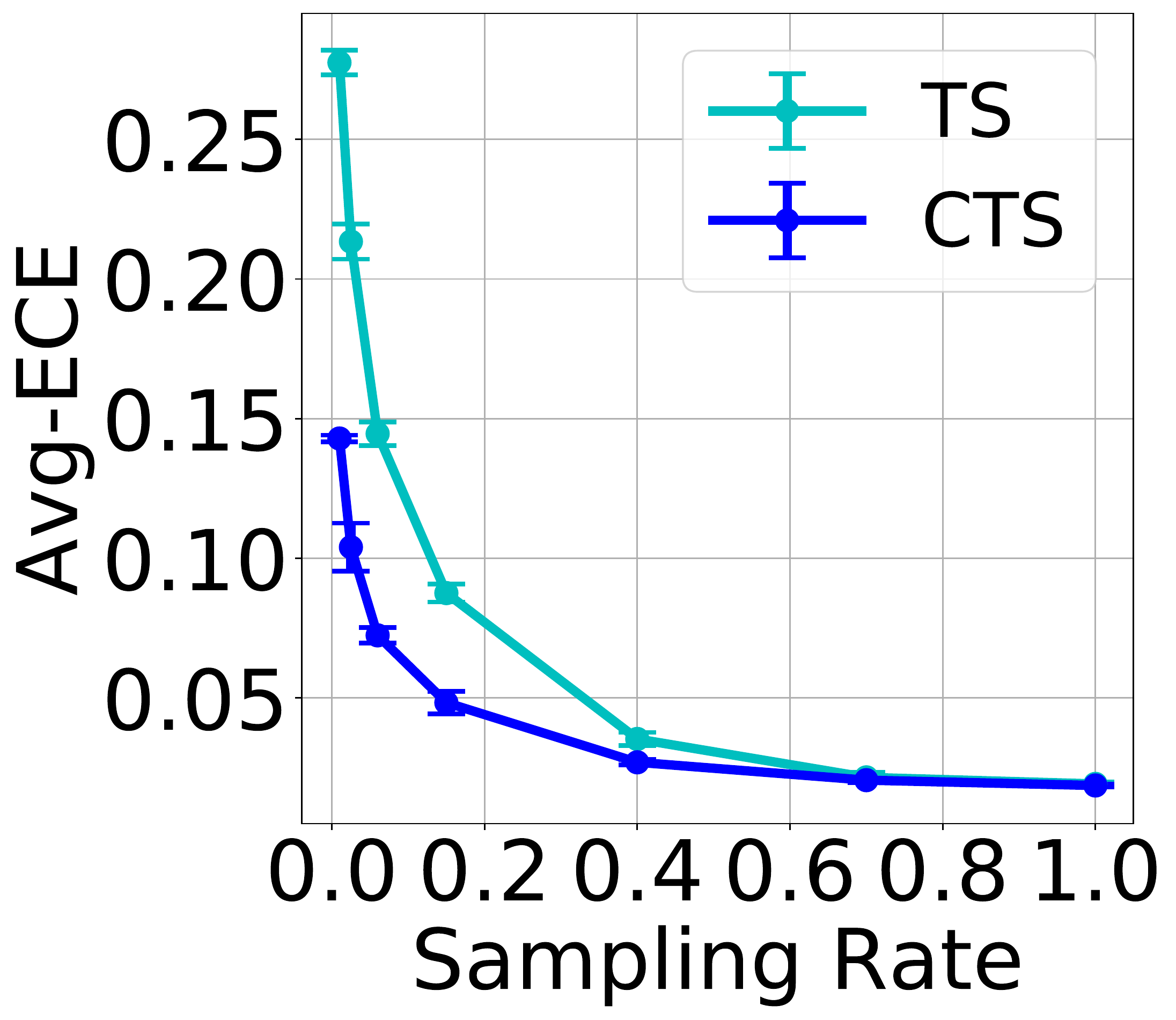}}
		\caption{Avg-ECE error}\label{fig unbalancec}
	\end{subfigure}	
	\caption{Sample imbalance: Calibration error as a function of the training set sampling rate.}\label{fig fig fig10}\vs\vs
\end{figure}\vs

\newpage

\section{Proofs of Main Theorems}

\subsection{Proof of Theorem \ref{thm under}}


Below is a restatement of Theorem \ref{thm under} where we provide separate statements for the sets $\Sc_1$ and $\Sc_2$.
\begin{theorem}\label{thm under2} There exists a distribution $\Dc$ (with unit $\ell_2$ norm inputs $\Xc$) as follows. Generate datasets $\Sc_1=(X_i,Y_i)_{i=1}^n\distas\Dc$ and $\Sc_2=(X_i,Y_i)_{i=1}^{30n}\distas\Dc$ and fix $R>0$. Minimize the empirical NLL loss to find linear classifiers $f_1,f_2$ as follows.
\[
f_i=\arg\min_{f\in \{f_{\ab,b}~|~\tn{\ab}\leq R\}} \text{NLL}(f,\Sc_i).
\]
Given precision $\eps>0$, we have the following accuracy and confidence behavior over $\Sc_1$ and $\Sc_2$.\vspace{-3pt}
\begin{myitemize}
\item Suppose $R\geq 3\log \max(6,\eps^{-1})$. Then, with probability at least $0.95-2\e^{-n/100}$, for all $(X,Y)\sim \Dc$ and $i\in\{1,2\}$: 
\[
\text{For all $X\in\Xc$:}~\hat{P}_{f_1}(X)\geq 1-\eps\quad\text{and}\quad \Pro_{(Y,X)\sim\Dc}(\hat{Y}_{f_1}(X)=Y)\leq 1-{1}/{20n}.
\]
\item Suppose $R\geq 6\log \max(50n,\eps^{-1})$ and $\eps<1/2$. Then, with probability at least $0.9$, 
\[
\text{For all $X\in\Xc$:}~\hat{P}_{f_2}(X)\geq 1-\eps\quad\text{and}\quad \Pro_{(Y,X)\sim\Dc}(\hat{Y}_{f_2}(X)=Y)= 1.
\]
\end{myitemize}\vspace{-3pt}
\end{theorem}
\begin{proof} The NLL (cross-entropy) loss on a dataset $\Sc$ is given by
\[
\text{NLL}(f_{\ab,b},\Sc)=-\frac{1}{n} \sum_{i=1}^n \log(\frac{\e^{y_i(\ab^TX_i+b)}}{1+\e^{y_i(\ab^TX_i+b)}})=\frac{1}{n} \sum_{i=1}^n \log({1+\e^{-y_i(\ab^TX_i+b)}})
\]
Fix orthogonal unit $\ell_2$ norm vectors $\ub,\vb\in\R^p$. Set $\vb'=(\ub+\vb)/\sqrt{2}$. Define the binary distribution $\Dc$ as follows.
\begin{align}
&\Pro(Y=1\bgl X=\vb)=\Pro(Y=0\bgl X=\vb')=\Pro(Y=0\bgl X=-\vb)=1\\
&\Pro(X=\vb)=1/2,\quad \Pro(X=\vb')=1/N,\quad \Pro(X=-\vb)=1/2-1/N.
\end{align}
Let $E_i$ be the event that $\vb'$ appears as an input in dataset $\Sc_i$. Observe that
\[
\e^{-n/N}\geq 1-\Pro(E_1)=(1-1/N)^n\geq 1-n/N.
\]
Thus, setting $N=20n$, we find $\Pro(E_1)\leq 0.05$ and $\Pro(E_2)\geq1- \e^{-50n/20n}=1-\e^{-2.5}\geq 0.91$. Also let $B$ be the event that at least $1/3$ of the training inputs are equal to $\vb$ and at least $1/3$ are equal to $-\vb$. Applying a standard Chernoff bound yields that $\Pro(B)\geq 1-2\e^{-\frac{n}{100}}$.

Before proceeding further, we also note that for all $x\geq 0$, we have
\[
\e^{-x}/2\leq \log(1+{\e^{-x}})\leq \e^{-x}
\]

\noindent{\bf{Analyzing $\Sc_1$ on the event $E_1\cap B$:}} Suppose $E_1$ holds. Note that the training dataset only contains inputs $\vb$ and $-\vb$. Thus, it can be concluded that the optimal classifier has the form $R\vb+b$ for some scalar $b$ i.e.~$\ab=R'\vb$ for $|R'|\leq R$. Let $0\leq\gamma\leq 1$ denote the fraction of $+\vb$ inputs within the training data. The empirical (training) NLL is given by
\[
\text{NLL}(f_{\ab,b},\Sc_1)=\gamma\log(1+\frac{1}{\e^{R'+b}})+(1-\gamma)\log(1+\frac{1}{\e^{R'-b}})
\]
Minimizing NLL over $R'$ reveals $R'=R$ and the loss is given by
\[
\text{NLL}(f_{\ab,b},\Sc_1)=\gamma\log(1+\frac{1}{\e^{R+b}})+(1-\gamma)\log(1+\frac{1}{\e^{R-b}})
\]
We next bound the optimal $b$ choice. Under event $B$, $\gamma,1-\gamma\geq 1/3$. Using $\ab=R\vb,b=0$ as an upper bound, we have that 
\begin{align}
\e^{-R}&\geq \gamma\log(1+\frac{1}{\e^{R+b}})+(1-\gamma)\log(1+\frac{1}{\e^{R-b}})\\
&\geq \frac{1}{3}\log(1+\frac{1}{\e^{R-|b|}})\geq \min(\frac{1}{6}\frac{1}{\e^{R-|b|}},\frac{\log(2)}{3})
\end{align}
which implies $|b|\leq \log 6$. Now observe that optimal classifier (on training), which obeys $\ab=R\vb,|b|\leq \log 6$, outputs the wrong decision on $\vb'$ since
\[
\hat{Y}_{f_1}(\vb')=\text{sign}(\ab^T\vb'+b)=\text{sign}(\frac{R}{\sqrt{2}}+|b|)=1
\]
as $R\geq \sqrt{2}\log 6$. This implies $\Pro(\hat{Y}_{f_1}(X)=Y)\leq 1-{1}/{20n}$. However, confidence on $\vb'$ (as well as on $\pm\vb$) is lower bounded as follows
\[
\hat{P}_{f_1}(X)\geq \frac{1}{1+\e^{-(R/\sqrt{2}-\log 6)}}\geq 1-\e^{-(R/\sqrt{2}-\log 6)}\geq 1-\eps
\]
whenever $R\geq \sqrt{2}(\log 6+\log(1/\eps))$ which is implied by $R\geq 3\log \max(6,1/\eps)$.

\noindent{\bf{Analyzing $\Sc_2$ on the event $E_2$:}} We claim that the classifier achieves small loss on all examples $\vb,-\vb,\vb'$ which will help show the result. First we pick a baseline classifier $\ab=R\frac{\vb-2\sqrt{2}\ub}{3}$ and $b=0$. This guarantees that for all $(Y,X)\sim \Dc$
\[
YX^T\ab \geq R/3.
\]
Thus empirical NLL over $\Sc_2$ is at most $-\log(\frac{\e^{R/3}}{1+\e^{R/3}})=\log(1+\frac{1}{\e^{R/3}})\leq \e^{-R/3}$. The overall loss will bound the individual losses i.e. at the optimal classifier $(\ab,b)$ (on training data) for any training example $(X,Y)\in\Sc_2$ we have
\[
\e^{-R/3}\geq \text{NLL}(f_{\ab,b},\Sc_2)\geq\frac{-1}{50n}(Y\log f(X)+(1-Y)\log(1-f(X)))
\]
Since $R\geq6\log (50n)$, we find $50n\e^{-R/3}\leq \e^{-R/6}$. Without losing generality, let us assume $Y=1$. This implies
\[
\e^{-R/6}\geq -\log f(X)\implies  f(X)\geq \e^{-\e^{-R/6}} \implies f(X)\geq 1-\e^{-R/6}.
\]
To achieve $1-\eps$ probability, we need $\e^{-R/6}\leq \eps$ which holds whenever $R\geq 6\log(1/\eps)$. For $\eps<1/2$, this also implies the classification is correct i.e. $Y=\hat{Y}$ since $f(X)>1/2$. The identical argument holds when $Y=0$.
%
\end{proof}


\subsection{Proof of Lemma \ref{thm simple}}
\begin{proof} Classifier outputs the probability $f(X)=\frac{\e^{\ab^TX+b}}{1+\e^{\ab^TX+b}}$. Note that $X=x\vb$ for $x\in\{-1,1\}$ hence without losing generality, we can assume $\ab=a\vb$ since any direction orthogonal to $\vb$ has zero inner product with input. Then, classifier simplifies to a single dimension as follows
\[
f(X)=\frac{\e^{ax+b}}{1+\e^{ax+b}}
\]
We need to find $a_*,b_*$ that maximizes the negative NLL loss
\[
-\Lc(\ab,b)=\E[Y\log(f(X))+(1-Y)\log(1-f(X))]
\]
This expectation leads to the scalar optimization
\begin{align*}
-2\Lc(\ab,b)=&(1-p_+)\log(\frac{\e^{a+b}}{1+\e^{a+b}})+p_+\log(\frac{1}{1+\e^{a+b}})+(1-p_-)\log(\frac{\e^{a-b}}{1+\e^{a-b}})+p_-\log(\frac{1}{1+\e^{a-b}}).
\end{align*}
Note that we can re-parameterize the loss by considering it as a function of $\alpha=a+b$ and $\beta=a-b$. Together it gives
\begin{align}
-2\Lc(\alpha,\beta)=&(1-p_+)\log(\frac{\e^{\alpha}}{1+\e^{\alpha}})+p_+\log(\frac{1}{1+\e^{\alpha}})+(1-p_-)\log(\frac{\e^{\beta}}{1+\e^{\beta}})+p_-\log(\frac{1}{1+\e^{\beta}}).\nn
\end{align}
Right hand side is maximized when partial derivatives with respect to $\alpha$ and $\beta$ are zero i.e.
\begin{align*}
\frac{-2\partial\Lc(\alpha,\beta)}{\partial \alpha}=&[(1-p_+)\frac{1}{1+\e^{\alpha}}-p_+\frac{1}{1+\e^{-\alpha}})]\\
\frac{-2\partial\Lc(\alpha,\beta)}{\partial \beta}=&[(1-p_-)\frac{1}{1+\e^{\beta}}-p_-\frac{1}{1+\e^{-\beta}}].
\end{align*}
Note that partial derivative w.r.t.~$\alpha$ depends only on $p_+$ and partial derivative w.r.t.~$\beta$ depends only on $p_-$ which greatly simplifies our life. Proceeding, we find that $\alpha_*=a_*+b_*$ satisfies the likelihood ratio
\begin{align}
&\frac{1-p_+}{1+\e^{\alpha_*}}-\frac{p_+}{1+\e^{-\alpha_*}}=0\implies \frac{1+\e^{\alpha_*}}{1+\e^{-\alpha_*}}=\frac{1-p_+}{p_+}
\end{align}
Note that this implies that the classifier output is
\[
f(\vb)=\frac{\e^{a_*+b_*}}{1+\e^{a_*+b_*}}=\frac{\e^{\alpha_*}}{1+\e^{\alpha_*}}=1-p_+.
\]
Similarly, following $\beta_*=a_*-b_*$, we find $\frac{1+\e^{\beta_*}}{1+\e^{-\beta_*}}=\frac{1-p_-}{p_-}$ and $f(-\vb)=p_-$. On the other hand, this classifier always predicts $1$ for $X=\vb$ and $0$ for $X=-\vb$ (as $1-p_+>1/2$ and $p_-<1/2$). As a result, since the test data is distributed with $\Dc_{\text{noisy}}(p_{\text{test}},p_{\text{test}})$, the test accuracy will be $1-p_{\text{test}}$ for both classes.
\end{proof}



\subsection{Proof of Theorem \ref{multitask thm}}


\begin{proof} Let us start with a change of variable from $\bal$ to $\bbeta$ where $\beta_0=\alpha_0$ and $\beta_i=\alpha_i-\alpha_0$. The goal is centering the $\alpha_i$ terms for simplifying subsequent discussion. Let $\Mc_{\bbeta}$ be the constraint set $\Mc$ adapted to the variable $\bbeta$ (which is a function of $\bal$). The proof will utilize a standard covering and union bound arguments. Fix covering resolution $\eps>0$ and let $S_0$ be an $\eps$ $\ell_2$-covering of the set $[\alpha_-,\alpha_+]$ and $S_i$ be an $\eps$ covering of the set $[-\Gamma,\Gamma]$. Clearly $|S_0|\leq 1\vee\frac{\alpha_+-\alpha_-}{\eps}$ and $|S_i|\leq1\vee \frac{2\Gamma}{\eps}$ where $\vee$ returns the maximum of two scalars. Let $\Mc'_{\bbeta}$ be a cover of $\Mc_{\bbeta}$ defined as
\[
\Mc'_{\bbeta}=\{\bbeta~\bgl ~\bbeta_i\in S_i\quad\text{and}\quad \bbeta_0\in S_0 \quad\forall i\geq 1\}.
\]
First let us bound the loss between $\bbeta\in\Mc_\bbeta$ and $\bbeta'\in\Mc'_\bbeta$ such that $\bbeta'$ is an $\eps$-neighbor of $\bbeta$ in the sense that $|\beta'_j-\beta_j|\leq \eps$ for $j\geq 0$. Define $\closs(\bal,\Dc,\hat{Y})=\E[\closs(Y,f_{\alpha_{\hat{Y}}}(X))\bgl \hat{Y}]$. Note that $\closs(\bal,\Dc,\hat{Y})$ is $L_fL_c$ Lipschitz function of $\alpha_{\hat{Y}}$ via initial assumptions and $1$-Lipschitzness of softmax \cite{gao2017properties}. Consequently
\begin{align}
|\closs(\bal,\Dc,\hat{Y})-\closs(\bal',\Dc,\hat{Y})|&\leq 2L_fL_c\eps.\label{d bound}
\end{align}
Taking expectation over $\hat{Y}$, this also implies
\begin{align}
|\closs(\bal,\Dc)-\closs(\bal',\Dc)|&\leq \E_{\hat{Y}}|\closs(\bal,\Dc,\hat{Y})-\closs(\bal',\Dc,\hat{Y})|\leq 2L_fL_c\eps.
\end{align}
We similarly have $|\closs(\bal,\Sc)-\closs(\bal',\Sc)|\leq 2L_fL_c\eps$. What remains is showing the closeness of finite sample and population over the cover which is a standard argument. Observe that each sample is independent and $\closs(Y,f_{\alpha_{\hat{Y}}}(X))$ is a random variable bounded by $1$. For each $\bbeta'\in\Mc'_{\bbeta}$, Bernstein bound guarantees that with probability at least $1-2\e^{-\frac{t^2}{2+t/\sqrt{n}}}$
\begin{align}
|\closs(\bal',\Dc)-\closs(\bal',\Sc)|\leq \frac{t}{\sqrt{n}}.\label{u bound}
\end{align}
Observe that cardinality of the cover obeys $|\Mc'|=|\Mc'_{\bbeta}|\leq (1\vee\frac{2\Gamma}{\eps})^K(1\vee\frac{\alpha_+-\alpha_-}{\eps})$. Pick $n\geq t^2=3\log |\Mc'|+3\log(2/\delta)$. Union bounding over the cover, this yields that, for all $\bbeta'\in\Mc'_{\bbeta}$, the bound above holds with probability at least
\[
1-2|\Mc'|\e^{-\frac{t^2}{2+t/\sqrt{n}}}\geq 1-2|\Mc'|\e^{-\frac{t^2}{3}}\geq 1-\delta.
\]
Overall, using triangle inequality and combining \eqref{d bound} and \eqref{u bound}, with $1-\delta$ probability, we obtain that, for all $\bal\in\Mc$, 
\begin{align}
|\closs(\bal,\Dc)-\closs(\bal,\Sc)|\leq  4L_fL_c\eps+\frac{\sqrt{3}\sqrt{K\log(1\vee\frac{2\Gamma}{\eps})+\log(1\vee\frac{\alpha_+-\alpha_-}{\eps})}}{\sqrt{n}}+\sqrt{\frac{3\log(2/\delta)}{n}}\label{main main main}.
\end{align}
To further simplify, set $\eps=\frac{1}{4\sqrt{n}L_fL_c}$. If $2\Gamma\geq\eps$, \eqref{main main main} yields
\begin{align}
|\closs(\bal,\Dc)-\closs(\bal,\Sc)|&\leq 2\sqrt{\frac{\log(\frac{\alpha_+-\alpha_-}{2\Gamma})+(K+1)\log({8\Gamma}{\sqrt{n}L_fL_c})}{{n}}}+4\sqrt{\frac{\log(2/\delta)}{n}}.\label{bound eps 1}
\end{align}
If $\eps>2\Gamma$, using $|\Mc'|\leq 1\vee \frac{\alpha_+-\alpha_-}{\eps}$, the bound \eqref{main main main} simplifies to 
\begin{align}
|\closs(\bal,\Dc)-\closs(\bal,\Sc)|\leq 2\sqrt{\frac{\log(1+4(\alpha_+-\alpha_-)\sqrt{n}L_fL_c)}{n}}+4\sqrt{\frac{\log(2/\delta)}{n}}.\label{exact state}
\end{align}
Since these hold for all $\bal\in\Mc$, using triangle inequality on minimizer of empirical loss $\bal^\star$ and the minimizer of population loss, we always have
\begin{align}
\closs(\bal^\star,\Dc)\leq \min_{\bal\in\Mc}\closs(\bal,\Dc)+2\sup_{\bal\in\Mc}|\closs(\bal,\Dc)-\closs(\bal,\Sc)|.
\end{align}
This concludes the proof after accounting for both cases (\eqref{bound eps 1} and \eqref{exact state} bounds) in the statement.
\end{proof}

\subsection{Proof of Corollary \ref{multitask cor}}

\begin{proof} The first step is ensuring that each sample set $\Sc_k$ with predicted labels $\hat{Y}=k$ has at least $p_{\min} n/2$ samples. Note that in expectation the cardinality obeys $\E|\Sc_k|\geq p_{\min}n$. Hence, applying a standard Chernoff bound followed by a union bound over $K$ classes gives an overall probability of success of at least
\begin{align}
\Pro(\bigcup_{k=1}^K \{|\Sc_k|\geq p_{\min}n/2\})\geq 1-K\exp(-\frac{p_{\min}n}{8}).\label{nk union bound}
\end{align}
The rest of the proof is by applying the result of Theorem \ref{multitask thm} on individual subsets $\Sc_k$ of the validation set. Each $\alpha_k$ solves the minimization
\[
\alpha_k^\star=\arg\min_{\alpha_-\leq \alpha_k\leq \alpha_+}\closs(\alpha_k,\Sc_k).
\]
For fixed $k$ (and $\Sc_k$), we can apply Theorem \ref{multitask thm} with $K\gets 1$, $\Gamma\gets 0$, $\Dc\gets \Dc_k$ to find that, with probability at least $1-\delta$
\begin{align}
\closs(\alpha_k^\star,\Dc_k)\leq \min_{\alpha_-\leq \alpha\leq \alpha_+}\closs(\alpha,\Dc_k)+4\sqrt{\frac{\log(1+4(\alpha_+-\alpha_-)\sqrt{n}L_fL_c)}{n_k}}+8\sqrt{\frac{\log(2/\delta)}{n_k}},\nn
\end{align}
where $n_k=|\Sc_k|$. Conditioning on $n_k\geq p_{\min}n/2$ via \eqref{nk union bound} and union bounding over all $1\leq k\leq K$, we conclude with the advertised bound of \eqref{ind bound}.
\end{proof}


%
%
%
%
%
%
%

\end{document}